\documentclass[manuscript,screen,authorversion,nonacm]{acmart} 
\AtBeginDocument{%
  }

\setcopyright{acmlicensed}
\copyrightyear{2027}
\acmYear{2027}
\acmDOI{XXXXXXX.XXXXXXX}
\acmConference[Conference acronym 'XX]{Make sure to enter the correct
  conference title from your rights confirmation email}{June 03--05,
  2018}{Woodstock, NY}
\acmISBN{978-1-4503-XXXX-X/2018/06}

\usepackage{multirow}
\usepackage{subcaption}

\newcommand{\E}{\mathbb E}
\newcommand{\Pp}{\mathbb P}
\newcommand{\Lcal}{\mathcal L}

\newcommand{\Ucal}{\mathcal U}
\newcommand{\Ucalj}
{\mathcal U_{\mathrm{joint}}}
\newcommand{\W}{W_2}

\newcommand{\indep}{\perp\!\!\!\perp}

\usepackage{xurl}
\usepackage{wasysym}
\usepackage{pifont}
\usepackage[misc]{ifsym}
\makeatletter
\def\@fnsymbol#1{%
   \ifcase#1\or
   \TextOrMath ~ \dagger\or
   \TextOrMath {\footnotesize\Letter} \dagger\or
   \TextOrMath \textdaggerdbl \ddagger \or
   \TextOrMath \textsection  \mathsection\or
   \TextOrMath \textparagraph \mathparagraph\or
   \TextOrMath \textbardbl \|\or
   \TextOrMath {\textdagger\textdagger}{\dagger\dagger}\or
   \TextOrMath {\textdaggerdbl\textdaggerdbl}{\ddagger\ddagger}\else
   \@ctrerr \fi
}
\renewcommand{\@mkauthorsaddresses}{}
\renewcommand{\footnotetextauthorsaddresses}[1]{}
\makeatletter
\renewcommand{\@titlefont}{\Huge\bfseries\sffamily\centering}

\renewcommand{\@mkauthors}{%
    \begingroup 
    \tiny
    \hsize=\textwidth
    \@mkauthors@iii
    \endgroup
}
\makeatother

\begin{document}

\title{Robbing Peter to Pay Paul: 
Cross-Attribute Externalities of Fairness Interventions}

\author{\normalsize{Agathe Fernandes Machado}{\textsuperscript{\Letter}}}
\begingroup
\renewcommand{\thefootnote}{}
\footnotetext{\Letter\ Corresponding author: fernandes\_machado.agathe@courrier.uqam.ca}
\endgroup
\orcid{0009-0002-9743-9708}
\affiliation{%
  \institution{\small Université du Québec à Montréal (UQÀM)}
  \city{Montréal}
  \state{Qc}
  \country{Canada}
}

\author{\normalsize Charlotte Jamotton}
\orcid{0009-0002-1227-3252}
\affiliation{%
  \institution{\small Université du Québec à Montréal (UQÀM)}
  \city{Montréal}
  \state{Qc}
  \country{Canada}
}

\author{\normalsize Arthur Charpentier}
\orcid{0000-0003-3654-6286}
\affiliation{%
  \institution{\small Université du Québec à Montréal (UQÀM)}
  \city{Montréal}
  \state{Qc}
  \country{Canada}
}

\renewcommand{\shortauthors}{Fernandes Machado et al.}

\begin{abstract}

Existing work on fairness interventions typically targets disparities associated with a single sensitive attribute in binary classification tasks. However, when multiple sensitive attributes are associated with a data subject, improving fairness with respect to one attribute does not necessarily improve fairness across all attributes: it may exacerbate disparities associated with other sensitive attributes, including those that are unavailable to the practitioner. We refer to these unintended effects as cross-attribute externalities of fairness. In this work, we study how fairness with respect to one sensitive attribute varies when an initially unfair classification or regression model is corrected with respect to another attribute. We characterize these externalities using Wasserstein barycentric disparity indices and develop a distributional framework that captures changes in both marginal and joint fairness measures. This perspective highlights a limitation of conventional fairness evaluation, in which the success of an intervention is assessed mainly for the attribute it explicitly targets. In a jointly Gaussian model, we derive explicit trajectories for marginal and joint disparities and identify conditions under which an intervention eliminates disparities with respect to its target attribute while increasing disparities associated with another attribute. We also derive sharp partial-identification bounds on the resulting externalities when one sensitive attribute is unavailable (e.g., race) and extend our results to categorical sensitive attributes. We illustrate these externalities using real-world classification and regression tasks: in the classification experiments, interventions increase unfairness with respect to the hidden attribute in the majority of scenarios across more than half of the datasets considered; in the regression experiments on the ACS Income dataset, correcting for race reduces sex disparities, whereas subsequently correcting for age increases them, highlighting how sequential fairness interventions can produce competing and non-monotonic effects across sensitive attributes.
\end{abstract}

\maketitle
\thispagestyle{empty}

\section{Introduction}


Machine learning (ML) systems are increasingly used to support decisions that affect access to employment, credit, education, and other essential resources \citep{Selbst2019, Buolamwini2018GenderSI, Scott2022}. Because these systems can reproduce disparities associated with protected characteristics \citep{Imana2024, Zilka2023, Arnold2021, Kchling2020}, a substantial body of research has developed methods \citep{Ashurst2023, Pessach2022, Mitchell2021} to measure and mitigate algorithmic unfairness. Yet existing fairness interventions are often designed and evaluated with respect to one sensitive attribute at a time, even though individuals may belong to multiple protected groups \citep{ gohar2023understanding}. This matters because reducing disparities associated with one attribute does not necessarily reduce disparities associated with another. In some cases, an intervention can improve fairness for one group while worsening it for another. Such cross-attribute externalities have been observed in empirical studies of fairness mitigation, including settings in which improvements for one protected attribute transfer residual disparities to another \citep{chen2024multiple,ai7080312,causalfairness2022}. These findings raise a basic question: when an intervention targets one sensitive attribute, what can we say about its effects on disparities associated with the other protected attributes?


This question is related to the challenges of fairness with multiple sensitive attributes (MSA). Although work has begun to extend fairness mitigation to MSA settings \citep{fairmask2023peng,causalfairness2022,monnier2026msa,msagraphs2026}, evaluating fairness across multiple attributes remains difficult. One reason is that marginal and intersectional disparities capture different aspects of a score distribution. A marginal audit compares the score distributions of groups defined by a single attribute, whereas an intersectional audit considers groups defined by combinations of attributes. Equal score distributions across the categories of each attribute separately do not imply equal distributions across their intersections \citep{molina2022bounding, Wang2022}. This is the distributional analogue of intersectional fairness and fairness gerrymandering \citep{kearns2018preventing,foulds2020intersectional,ghosh2021characterizing,herlihy2024structured}, which shows how disparities can persist within subgroups even when aggregate measures suggest parity. Consequently, assessing the effect of a fairness intervention requires more than checking whether the disparity it was designed to reduce has decreased.

We assess this problem through a distributional framework based on Wasserstein distances. Rather than evaluating fairness through a single summary of group outcomes, we compare the entire conditional score distributions using Wasserstein barycentric disparity indices, thereby assessing fairness under the demographic parity (DP) criterion for scores produced by an ML model \cite{barocas2023fairness}. These indices quantify the dispersion of group-conditional distributions around their Wasserstein barycenter and can be defined for both marginal and joint sensitive profiles \citep{legouic2020projection}. We then study a rank-preserving barycentric intervention that moves scores toward the barycenter associated with the target sensitive attribute. The correction has a useful mathematical property: the targeted disparity contracts exactly at a quadratic rate as the correction intensity increases. However, this guarantee applies only to the targeted marginal audit. Considering two sensitive attributes, the effects on the remaining marginal disparity and on the joint disparity are not determined by the same contraction result. We therefore define a directed externality as the change in a non-targeted disparity index induced by the correction. This provides a common framework for examining which disparities decrease, which increase, and how these changes depend on the underlying score distributions.

These questions are complicated further when some sensitive attributes are unobserved or unavailable for auditing. Protected characteristics may be uncollected or incompletely observed because of legal restrictions, privacy concerns, or institutional practice \cite{discriminationlaw2012eu,haley2022collection}. One possible response is to infer missing attributes from auxiliary information or weak proxies \citep{infersa2025wang, pmlr-v202-zhu23n}. However, such inference does not generally recover group membership with certainty, and its use introduces additional assumptions in the audit. Here, we consider a complementary question: what can be established about cross-attribute externalities from the information that is available? Under Gaussian assumptions with two sensitive attributes, we derive sharp bounds on the change of disparity associated with an unobserved attribute when the relevant correlations are only partially known. We extend this partial-identification analysis to categorical unobserved attributes, characterizing the exact identified set of externalities compatible with the observed score distributions and known group proportions. The objective is not to reconstruct an unknown attribute exactly, but to characterize the range of distributional consequences (e.g., how limited information about the sensitive profile can leave the sign or magnitude of a cross-attribute externality unidentified). 


\subsection{Related Work}

The motivation for the present analysis is supported by empirical evidence that fairness interventions may shift disparities across protected attributes (i.e., generate cross-attribute externalities), but it addresses a different question from studies that compare the performance of particular mitigation algorithms. For instance, \citep{chen2024multiple} evaluate mitigation methods using DP and equalized odds for classification tasks across intersectional groups and report that improving fairness for one attribute can worsen fairness for unconsidered attributes; in their Adult-dataset experiments involving sex and race, mitigating unfairness with respect to sex worsens fairness with respect to the unconsidered attribute race in 55\% of the scenarios considered. Related effects have been reported in credit scoring, where methods targeting one protected attribute at a time have been observed to transfer residual harm to other monitored attribute \citep{ai7080312}, and in experiments on the Adult-dataset where reducing gender-related unfairness can increase race-related unfairness \citep{causalfairness2022}. These findings establish the practical relevance of cross-attribute externalities. Our contribution is complementary: rather than focusing on comparing mitigation algorithms on fairness criteria, we characterize how distributional Wasserstein disparities evolve under a specified intervention, identify conditions under which externalities can be adverse, and determine what can be learned when part of the sensitive profile is unobserved. We also examine these effects empirically across classification and regression tasks, considering both observed and unobserved sensitive attributes.


Our main contributions and findings are:
\begin{itemize}
    \item We characterize a rank-preserving barycentric correction and prove that it reduces the targeted disparity according to an exact quadratic contraction law. 
    \item In the case of two sensitive attributes, we define a directed externality vector to track changes in the remaining marginal and joint disparity indices. Based on this vector, we derive closed-form expressions for the marginal and joint disparity trajectories under a jointly Gaussian model and characterize the conditions under which a correction increases disparity with respect to a non-targeted attribute.
    
    \item When one protected attribute is observed and targeted by an intervention, we derive attainable bounds on cross-attribute externalities associated with an unobserved sensitive attribute under a Gaussian model, assuming that only selected correlations are known. For categorical attributes, we characterize the exact identified set of externalities compatible with the observed score distributions and known group proportions, and derive sharp empirical bounds when hidden-group counts are fixed.
    \item We examine the effects of Wasserstein barycentric correction on non-targeted disparities across multiple datasets, prediction tasks, and sensitive-attribute configurations. We extend the theoretical results from the two-attribute to the three-attribute setting, including scenarios in which one attribute is unavailable for auditing, and consider regression tasks.
\end{itemize}

The remainder of the paper is structured as follows: Section \ref{sec:wasserstein-audits} introduces the marginal and joint Wasserstein indices and studies the effects of targeted correction, including the Gaussian and interaction-based examples; Section \ref{sec:hidden-attribute-bounds} develops the partial-identification results for unobserved sensitive attributes; Section \ref{sec:results} evaluates the distributional effects of correction on real datasets. Proofs and additional results are provided in the Appendix.



\section{Wasserstein Fairness Audits for Multiple Sensitive Attributes}
\label{sec:wasserstein-audits}


In this Section, we develop the framework to assess the distributional consequences of a targeted correction without assuming that improvements in one audit imply improvements in the others. In the following, we first define barycentric indices for marginal and joint groups. Then, we consider an intervention that corrects the score distribution across groups defined by one sensitive attribute, characterizing its effect on the targeted disparity and introducing a directed externality vector to quantify changes in the remaining marginal and joint disparities. Through Gaussian and categorical examples, we define what marginal and joint measures can conceal, illustrating why neither marginal parity nor a reduction in joint disparity alone provides a complete account of the effects of a correction.

\subsection{Wasserstein Fairness Indices}
\label{subsec:notations}

Let $Z$ denote the real-valued score produced by a predictive model, for either a classification or a regression task. We assume that $\mathbb E[Z^2]<\infty$. Let $D=(D_1,D_2)$ denote a pair of categorical sensitive attributes, taking values in finite sets $\mathcal D_1$ and $\mathcal D_2$, respectively. For $j\in\{1,2\}$ and $a\in\mathcal{D}_j$, let $p_{j,a}=\Pp(D_j=a)$ and denote by $Q_{j,a}:[0,1]\to\mathbb R$ the quantile function of the conditional score distribution $\Lcal(Z\mid D_j=a)$. We quantify the marginal distributional disparity associated with each sensitive attribute using the corresponding barycentric index~\citep{legouic2020projection} 
\begin{equation}\label{eq:Um}
\begin{aligned}
\Ucal_j(Z) &= \inf_{\nu \in \mathcal{P}_2 (\mathbb{R})} \sum_{a \in \mathcal{D}_j} p_{j,a}\W^2 \bigl(\Lcal(Z\mid D_j=a),\nu \bigr) =\sum_{a \in \mathcal{D}_j} p_{j,a}\int_0^1 \bigl(Q_{j,a}(u)-\bar Q_j(u) \bigr)^2du,
\quad \bar Q_j(u)=\sum_{a \in \mathcal{D}_j} p_{j,a}Q_{j,a}(u),
\end{aligned}
\end{equation}
where $\mathcal{P}_2 (\mathbb{R})$ denotes the set of probability measures on $\mathbb{R}$ with finite second moments. The functional $\Ucal_j(Z)$ quantifies the weighted dispersion of the group-conditional score distributions around their Wasserstein barycenter. It is nonnegative and vanishes if and only if $\mathcal L(Z\mid D_j=a)=\mathcal L(Z\mid D_j=a')$ for every $a,a'\in\mathcal D_j$ with positive probability.

We extend this construction to joint sensitive profiles. For each $D=(a,b) \in \mathcal{D}_1 \times \mathcal{D}_2$ with $p_{ab} = \mathbb P (D_1=a,D_2=b)$, let $Q_{ab}$ denote the quantile function of the joint conditional law $\mathcal L (Z \mid D_1=a, D_2=b)$. The corresponding joint barycentric index is 
\begin{equation}\label{eq:Ujoint}
\Ucalj(Z)=\sum_{(a,b): \; p_{ab}>0} p_{ab} \int_0^1 \bigl(Q_{ab}(u)-\bar Q(u) \bigr)^2du,
\qquad \bar Q(u):=\sum_{(a,b): \; p_{ab}>0}p_{ab}Q_{ab}(u).
\end{equation}
In particular, the index $\Ucalj(Z)=0$ if and only if the conditional score distributions are identical across all positive-probability joint sensitive profiles.

The distinction between marginal and joint audits is structural. For any $a\in \mathcal{D}_1$ with $p_{1,a}>0$, the conditional law entering the marginal audit can be written as a mixture of joint-profile conditional laws
\begin{equation}
    \mathcal L(Z\mid D_1=a) = \sum_{b:\;p_{ab}>0} \; p_{ab}/p_{1,a} \; \mathcal L(Z\mid D_1=a, D_2=b) .
\end{equation}
Because quantiles do not, in general, commute with mixture formation, the marginal and joint barycentric indices do not admit a general additive decomposition: in general, $\Ucalj(Z) \ne \sum_j \Ucal_j(Z)$. 


\subsection{Wasserstein Barycentric Correction of a Single Sensitive Attribute}
\label{subsec:three-audits-one-intervention}

When multiple sensitive attributes are considered, correcting disparities with respect to one attribute may also affect disparities associated with the other. We now consider a barycentric correction targeting a single sensitive attribute and characterize its spillover effects on the marginal and joint fairness indices defined above in \eqref{eq:Um} and \eqref{eq:Ujoint}.

Consider a marginal distributional intervention targeting $D_1$. For each $a\in\mathcal D_1$ with $p_{1,a}>0$, let $F_{1,a}$ denote the conditional cumulative distribution function (CDF) of $\mathcal L(Z\mid D_1=a)$, and define the barycenter quantile function $\bar Q_1 = \sum_{a \in \mathcal{D}_1} p_{1,a} Q_{1,a}$. The transport map $T_1$ and the associated $\lambda$-interpolated score $Z_\lambda$ are given by
\begin{equation}\label{eq:repair}
T_1(z,a)=\bar Q_1(F_{1,a}(z)),\qquad Z_\lambda=(1-\lambda)Z+\lambda T_1(Z,D_1),\quad 0\le\lambda\le1.
\end{equation} 
Assuming continuous conditional score distributions, this construction maps each score to the barycenter quantile at the same conditional rank. In the presence of atoms, the transport can instead be defined using randomized conditional ranks. This is a distributional transformation of the score $Z$, not a causal intervention on $D_1$.

\begin{proposition}[Target correction]\label{prop:target}
For every $\lambda\in[0,1]$, $\Ucal_1(Z_\lambda)=(1-\lambda)^2 \; \Ucal_1(Z)$. In particular, $\Ucal_1(Z_1)=0$. The proof is given in Appendix \ref{app:subsec:proof-prop-2-1}.
\end{proposition}
The novel contribution is the \emph{directed externality vector} associated with the intervention $T_1$ at intensity $\lambda$
\begin{equation}\label{eq:externalities}
\mathcal E_{1}(\lambda)=\bigl(\Ucal_1(Z_\lambda)-\Ucal_1(Z),\;
\Ucal_2(Z_\lambda)-\Ucal_2(Z),\;\Ucalj(Z_\lambda)-\Ucalj(Z)\bigr).
\end{equation}
The first component is known exactly: $\mathcal E_{1,1}(\lambda) = -\lambda(2-\lambda)\mathcal U_1(Z) \leq 0$. 
The remaining components, $\mathcal E_{1,2}(\lambda)$ and $\mathcal E_{1,3}(\lambda)$, quantify the changes in marginal disparity, with respect to $D_2$, and in joint-profile disparity, respectively. Specifically $\mathcal E_{1,2}(\lambda)$ quantifies the distributional spillover on $D_2$ induced by correcting $D_1$, which need not follow the direction of change in the targeted audit. 

\begin{proposition}[Joint index dominates marginal indices]\label{prop:jointbound}
For categorical attributes and squared-integrable laws,
$\Ucalj(Z)\ge\max\{\Ucal_1(Z),\Ucal_2(Z)\}$. However, $\sum_j \Ucal_j$ do not, in general, give an additive decomposition of $\Ucalj$. The joint index dominates each marginal index in level, but this ordering does not extend to changes under correction. In particular, a reduction in a marginal disparity does not, by itself, imply a reduction in the joint disparity, or vice versa. The proof is given in Appendix \ref{app:subsec:proof-prop-2-2}.
\end{proposition}


\subsection{An Explicitly Solvable Gaussian-Attribute Example}
\label{subsec:joint-gaussian-ex}

To illustrate the distributional effects of a targeted correction and characterize when it can increase disparities in a non-targeted marginal audit, we consider an additive score with two jointly Gaussian sensitive attributes. In Appendix \ref{app:subsec:three_sensitive}, we extend it to three Gaussian sensitive attributes. This setting admits explicit expressions for the marginal and joint disparity indices along the correction trajectory, without requiring the score itself to be Gaussian. These expressions identify the parameter regimes in which the correction improves the targeted and joint audits but worsens the non-targeted marginal audit.

Let $(D_1,D_2)$ be centered, standard Gaussian random variables with correlation $\rho\in(-1,1)$. Consider an additive score of the form
\begin{equation}\label{eq:gauss-predictors}
Z=b_1D_1+b_2D_2+\boldsymbol{\theta}^\top\boldsymbol{X},
\end{equation}
where $\boldsymbol{X}\in\mathbb R^d$ contains additional explanatory variables. We take $\boldsymbol{X}$ to be centered and square-integrable. Projecting $\boldsymbol{\theta}^\top\boldsymbol{X}$ linearly onto $(D_1,D_2)$ gives
\[
\boldsymbol{\theta}^\top\boldsymbol{X}
=\kappa_1D_1+\kappa_2D_2+R_{\boldsymbol D},
\qquad
\begin{pmatrix}\kappa_1\\\kappa_2\end{pmatrix}
=\operatorname{Cov}(\boldsymbol D)^{-1}
  \operatorname{Cov}(\boldsymbol D,\boldsymbol{\theta}^\top\boldsymbol X),
\quad \boldsymbol D=(D_1,D_2)^\top.
\]
The projection residual is centered and orthogonal to $(D_1,D_2)$, but orthogonality alone does not imply independence. We additionally assume that $R_{\boldsymbol D}$ is independent of $(D_1,D_2)$; this holds, for example, if $(\boldsymbol{\theta}^\top\boldsymbol X,D_1,D_2)$ is jointly Gaussian. With $\beta_j=b_j+\kappa_j$, we thus obtain
\begin{equation}\label{eq:gauss}
Z=\beta_1D_1+\beta_2D_2+R_{\boldsymbol D},
\qquad
\mathbb E[R_{\boldsymbol D}]=0,\quad
\mathbb E[R_{\boldsymbol D}^2]<\infty,\quad
R_{\boldsymbol D}\perp(D_1,D_2).
\end{equation}
The residual need not be Gaussian. Moreover, $\beta_1$ and $\beta_2$ are coefficients of a statistical representation: they do not necessarily imply that both sensitive attributes are explicitly used to compute the score. In particular, $\beta_2$ may be nonzero even if $b_2=0$ in \eqref{eq:gauss-predictors}.

For this continuous-attribute setting, we use the integrated barycentric indices introduced above. Define $c_j=\operatorname{Cov}(Z,D_j)$, so that $c_1=\beta_1+\rho\beta_2$ and $c_2=\beta_2+\rho\beta_1$. Since $D_2=\rho D_1+\sqrt{1-\rho^2}\,G$ for a standard Gaussian $G$ independent of $(D_1,R_{\boldsymbol D})$, we have
\[
Z=c_1D_1+V_1,\qquad
V_1=\beta_2\sqrt{1-\rho^2}\,G+R_{\boldsymbol D},
\qquad V_1\perp D_1.
\]
Hence the conditional distributions $\mathcal L(Z\mid D_1=d)$ are translations of a common law, even if $Z$ is not Gaussian. Their Wasserstein barycenter is the law of $V_1$, and the barycentric repair targeting $D_1$ is therefore exactly
\begin{equation}\label{eq:gaussrepair}
Z_\lambda=Z-\lambda c_1D_1,
\qquad \lambda\in[0,1].
\end{equation}
At $\lambda=0$, the original score is recovered; at $\lambda=1$, the conditional score distributions given $D_1$ coincide, so that the targeted disparity vanishes.

\begin{proposition}[Three explicit disparity trajectories under Gaussian-attribute correction]\label{prop:gauss}
Under \eqref{eq:gauss}, the marginal and joint disparity indices evolve along the repair trajectory as
\begin{equation}\label{eq:paths}
\begin{aligned}
&\Ucal_1(Z_\lambda)
=(1-\lambda)^2c_1^2,&
\Ucal_2(Z_\lambda)
=(c_2-\lambda\rho c_1)^2,&
&\Ucalj(Z_\lambda)
=\beta_1^2+\beta_2^2+2\rho\beta_1\beta_2
-\lambda(2-\lambda)c_1^2.
\end{aligned}
\end{equation}
Moreover, the externality on the non-targeted marginal index, defined in \eqref{eq:externalities}, is
\begin{equation}\label{eq:gaussian_externality}
\mathcal E_{1,2}(\lambda)
=\lambda\rho c_1\bigl(\lambda\rho c_1-2c_2\bigr).
\end{equation}
The proof is given in Appendix \ref{app:subsec:proof-prop-2-3}.
\end{proposition}

\paragraph{Distinct behaviors of the disparity trajectories.}
The three indices follow different trajectories: the targeted disparity $\Ucal_1$ decreases quadratically to zero; the joint disparity $\Ucalj$ is non-increasing throughout the correction with $\lambda$; the non-targeted marginal disparity $\Ucal_2$ follows a quadratic trajectory whose behavior depends on the relative signs and magnitudes of $c_2$ and $\rho c_1$: it may increase from the outset or initially decrease before eventually exceeding its original value. At full correction ($\lambda=1$), the targeted disparity vanishes, while the non-targeted marginal disparity becomes $\Ucal_2(Z_1)=\beta_2^2(1-\rho^2)^2$, compared with its initial value $\Ucal_2(Z)=c_2^2=(\beta_2+\rho\beta_1)^2$. Thus, the externality $\mathcal E_{1,2}(1)$ is positive (i.e., full correction increases the non-targeted disparity) if and only if
\begin{equation}\label{eq:region}
|\beta_2|(1-\rho^2)>|\beta_2+\rho\beta_1|.
\end{equation}
This illustrates a form of \emph{masking and unmasking}: the initial marginal disparity can be small because the contributions associated with correlated variables partially cancel. Correcting the score with respect to $D_1$ can disrupt this cancellation, increasing the disparity measured with respect to $D_2$.

\paragraph{Explicit characterization of the deterioration regimes.}
The condition in \eqref{eq:region} admits an explicit characterization when $\beta_1>0$ and $0<\rho<1$. Writing $r=\beta_2/\beta_1$, full correction increases $\Ucal_2$ precisely when
\[
-\frac{1}{\rho}<r<-\frac{\rho}{2-\rho^2}.
\]
Within this interval, the onset of deterioration depends on $r$. Deterioration is \emph{immediate} when $-1/\rho<r\leq-\rho$: $\Ucal_2(Z_\lambda)>\Ucal_2(Z)$ for every $\lambda>0$. It is \emph{delayed} when $-\rho<r<-\rho/(2-\rho^2)$: $\Ucal_2$ initially decreases, then increases and eventually exceeds its original value. At $r=-\rho/(2-\rho^2)$, the initial and fully corrected disparities are equal.

\begin{figure}[t]
\centering\includegraphics[width=0.9\linewidth]{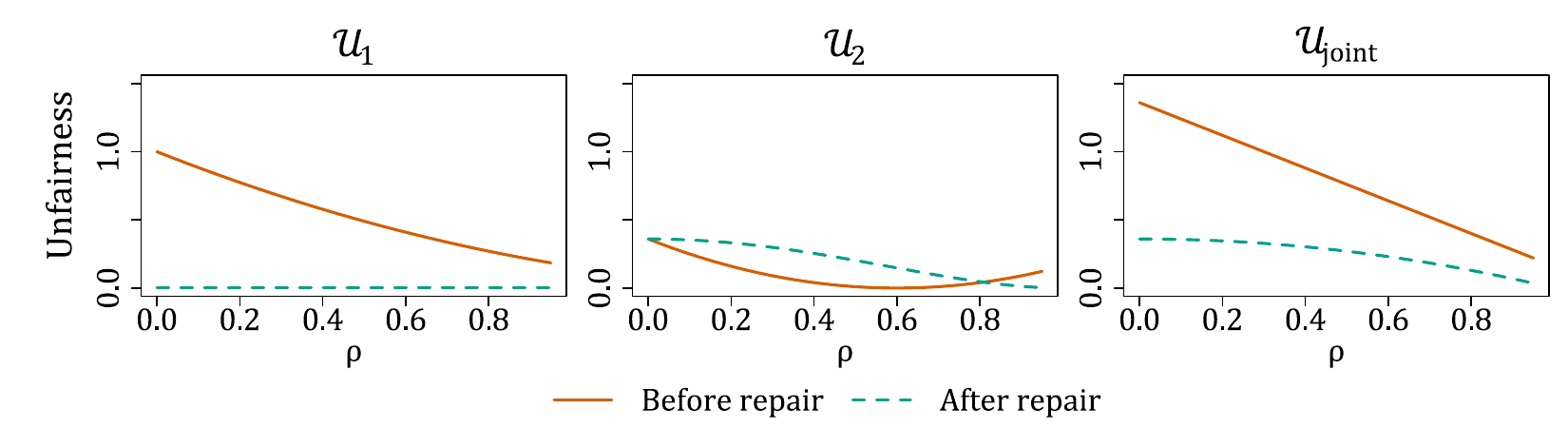}
\caption{\textbf{A before-and-after audit of full correction for $D_1$.} Gaussian sensitive attributes with $\beta_1=1$ and $\beta_2=-0.6$. The three disparity indices (the targeted disparity $\Ucal_1$, the cross-attribute externality $\Ucal_2$, and the joint unfairness $\Ucalj$) are plotted as functions of $\rho$. Solid curves: original score; dashed: full correction for $D_1$. The indices do not depend on the distribution of $R_{\boldsymbol D}$.}
\label{fig:one}
\end{figure}

\begin{figure}[ht]
\centering\includegraphics[width=0.9\linewidth]{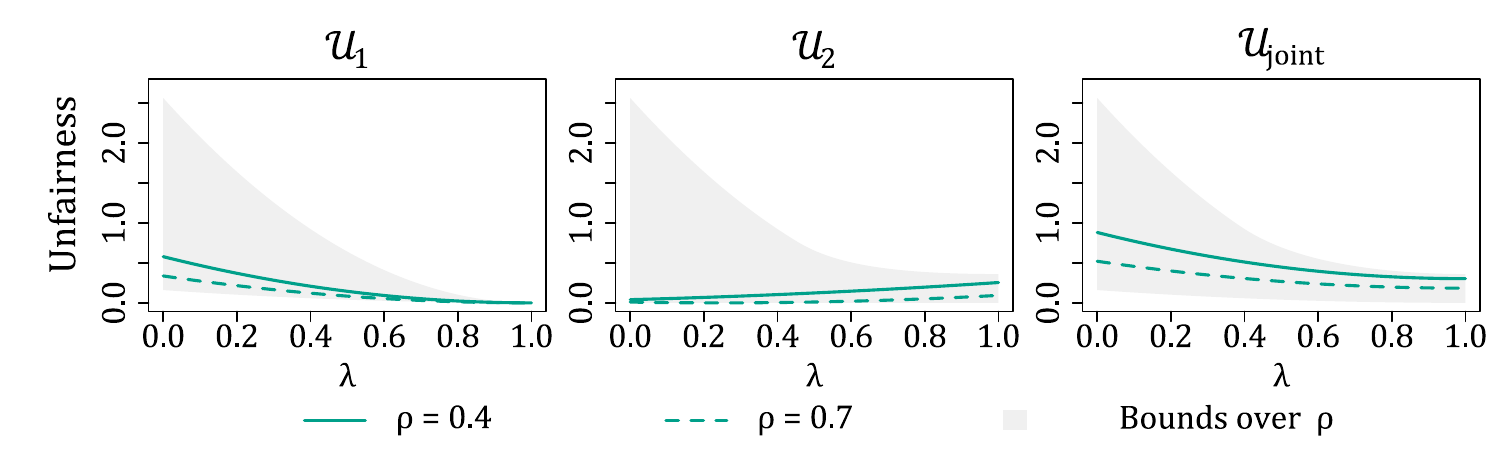}
\caption{\textbf{Partial correction need not have the same effects on different audits.} For $\rho \in \{0.4,0.7\}$, $\beta_1=1$, $\beta_2=-0.6$, the targeted ($\Ucal_1$) and joint ($\Ucalj$) disparities decrease, while $\Ucal_2$ immediately increases ($\rho=0.4$) or ends up increasing after an initial improvement ($\rho=0.7$). The exact trajectories follow \eqref{eq:paths}.}
\label{fig:paths}
\end{figure}

Figure~\ref{fig:one} compares the original and fully corrected disparity indices across values of $\rho$. The targeted index vanishes after full correction, but the non-targeted marginal index can increase. Figure~\ref{fig:paths} illustrates these distinct trajectories for two values of $\rho$, with $\beta_1=1$ and $\beta_2=-0.6$. In both cases, the targeted and joint disparities decrease throughout the correction. When $\rho=0.4$, the non-targeted disparity increases immediately. When $\rho=0.7$ and $r=-0.6$, it initially decreases, reaches its minimum at $\lambda\simeq0.246$, returns to its baseline level at approximately $\lambda\simeq0.493$, but eventually exceeds its original value. Thus, a correction can initially improve the non-targeted audit and later worsen it beyond its baseline level.

\paragraph{Correction dynamics.}
Setting $a=\rho c_1$, we have
$\Ucal_2(Z_\lambda)-\Ucal_2(Z)=\lambda a(\lambda a-2c_2)$.
If $a\neq0$, the quadratic trajectory has its unconstrained minimum at $\lambda^\star=c_2/a$ and returns to the initial level at $\lambda=2c_2/a$, whenever these values lie in $[0,1]$. Depending on their location, deterioration may occur immediately, appear only beyond a threshold, or never occur. This allows us to construct a phase diagram in the plane $(\rho,\beta_2/\beta_1)$ analytically, without simulation.

\subsection{What the Joint Measure Can Conceal}
\label{subsec:joint-measure-conceal}


\subsubsection{Gaussian Attributes} 

Under full correction in the Gaussian model,
\begin{equation}\label{eq:residual}
\Ucalj(Z_1)=\beta_2^2(1-\rho^2),\qquad
\Ucal_2(Z_1)=\beta_2^2(1-\rho^2)^2.
\end{equation}
Although the correction eliminates $\Ucal_1(Z_1)=0$, the joint index remains positive $\Ucalj(Z_1)>\Ucal_2(Z_1)$ whenever $\beta_2\rho\ne0$ and $|\rho|<1$. The full sensitive profile therefore retains information about the score beyond that captured by either marginal audit considered separately. A decrease in $\Ucalj$ does not, however, imply a decrease in $\Ucal_2$.

Table~\ref{tab:unfairness_rho} reports the initial and fully corrected indices for several values of $\rho$. The non-targeted disparity increases for some correlation values but decreases for others, even though the joint index decreases throughout the correction.
\begin{table}[t]
\centering
\caption{\textbf{Cross-attribute externalities under full correction for $D_1$.} Marginal ($\Ucal_1, \Ucal_2$) and joint ($\Ucalj$) disparity indices before and after full correction for $D_1$ for different values of $\rho\in[0,0.9]$ in a jointly Gaussian model with $(\beta_1,\beta_2)=(1,-0.6)$.}
\label{tab:unfairness_rho}
\begin{tabular}{c r r r r r}
\toprule
$\rho$ & $\Ucal_1(0)$ & $\Ucal_2(0)$ & $\Ucal_2(1)$ & $\Ucalj(0)$ & $\Ucalj(1)$\\
\midrule
$0.2$ & $0.7744$ & $0.1600$ & $0.3318$ & $1.1200$ & $0.3456$\\
$0.4$ & $0.5776$ & $0.0400$ & $0.2540$ & $0.8800$ & $0.3024$\\
$0.6$ & $0.4096$ & $0.0000$ & $0.1475$ & $0.6400$ & $0.2304$\\
$0.9$ & $0.2116$ & $0.0900$ & $0.0130$ & $0.2800$ & $0.0684$\\
\bottomrule
\end{tabular}
\end{table}

\paragraph{Attribution is not intervention.} For two attributes, consider the Shapley attribution induced by $v(\{1\})=\Ucal_1$, $v(\{2\})=\Ucal_2$, and $v(\{1,2\})=\Ucalj$. The resulting attribution to $D_1$ is $\phi_1=\tfrac12(\Ucal_1+\Ucalj-\Ucal_2)$. At $\rho=0.6$ in the Gaussian example, $\phi_1/\Ucalj=0.82$, whereas the proportion of the original joint disparity removed by full correction is $0.4096/0.64=0.64$. The difference illustrates that an attribution of disparity to an attribute need not equal the reduction achieved by correcting that attribute. The two quantities answer different questions: the Shapley value allocates disparity according to a specified characteristic function, whereas the second measures the effect of a particular correction.

The role of attribute ordering in sequential Wasserstein
audits and the corresponding order-independent Shapley
allocation are studied in
\cite{fernandesmachado2026aistats}.
Unlike these descriptive decompositions, our focus here
is on the distributional consequences of fairness
interventions.

\subsubsection{Categorical Attributes} 

We now return to categorical sensitive attributes. Let $D_1,D_2$ be independent and equiprobable on $\{-1,+1\}$, and consider the score
\begin{equation}\label{eq:xor}
Z=\alpha D_1+\kappa D_1D_2+R_{\boldsymbol{D}},
\qquad R_{\boldsymbol{D}} \sim\mathcal N(0,\sigma^2),\quad R_{\boldsymbol{D}}\perp \boldsymbol{D}.
\end{equation}

\paragraph{Two zero marginal audits, one positive joint audit.} Conditional on $D_1$, the score distributions differ only by a location shift. The full barycentric correction targeting $D_1$  therefore gives $Z_1=Z-\alpha D_1=\kappa D_1D_2+R_{\boldsymbol{D}}$. Since $D_1$ and $D_2$ are independent and equiprobable, the conditional distribution of $Z_1$ given either attribute alone is the same across its two groups. In contrast, the four joint profiles have conditional means $\pm\kappa$. It follows that
\begin{equation}\label{eq:xorresult}
\Ucal_1(Z_1)=\Ucal_2(Z_1)=0,\qquad \Ucalj(Z_1)=\kappa^2>0.
\end{equation}
For $\kappa\neq0$, both marginal audits indicate distributional parity, even though the four intersectional score distributions remain distinct. 

This example isolates an interaction effect that is invisible to the two marginal indices. It complements the Gaussian \emph{masking-and-unmasking} example: marginal and joint audits capture different aspects of distributional disparity, and neither the sum of the marginal indices nor the examination of each attribute can, in general, replace a joint audit.


\section{Sharp Bounds on Externalities for an Unavailable Attribute}
\label{sec:hidden-attribute-bounds}

We illustrate Section~\ref{sec:wasserstein-audits} in a setting where regulations prohibit discrimination based on MSA \cite{foulds2020intersectional, kearns2018preventing, morina2019auditing}, while privacy or legal restrictions prevent the collection of some attributes \cite{discriminationlaw2012eu, haley2022collection}. We consider two sensitive attributes (\(D_1,D_2\)), where decision-makers correct only for \(D_1\) (e.g., gender), while \(D_2\) (e.g., race) is unavailable for correction. Under the statistical representation \eqref{eq:gaussresidual-x}, \(D_2\) may be either unavailable when constructing \(Z\) (as formalized in Appendix \ref{app:subsec:joint-gaussian-ex-H} for \eqref{eq:gauss-predictors}) or it may be available for prediction but not correction. Even when excluded from \(Z\), disparities associated with \(D_2\) may still arise through proxies and statistical dependence on \(D_1\). Decision-makers may infer \(D_2\) from proxies and mitigate \(Z\) with respect to \((D_1,D_2)\), or correct for \(D_1\) alone, potentially exacerbating disparities associated with \(D_2\), as illustrated by \eqref{eq:region}. Assuming an external demographic source provides information on the dependence between \(D_1\) and \(D_2\), we ask: {Can the impact of correcting only for \(D_1\) on disparities associated with \(D_2\) be bounded?}

\subsection{Closed-Form Bounds for the Joint Gaussian Case}

Throughout this Section, we assume that $(D_1,D_2,Z)$ are centered, standardized Gaussian variables, and that the indices $\Ucal_j$ retain their earlier interpretation as Wasserstein barycentric disparities, as defined in Section~\ref{sec:wasserstein-audits}.

\begin{proposition}[Sharp bounds for a hidden externality]
\label{prop:hidden-sharp-fr}
Let $(D_1,D_2,Z)$ be a centered Gaussian vector with unit marginal variances. Suppose the correlations
\[
a=\operatorname{Corr}(D_1,Z),
\qquad \rho =\operatorname{Corr}(D_1,D_2),
\qquad |a|<1,\quad |\rho|<1,
\]
are known, while $b=\operatorname{Corr}(D_2,Z)$ is unknown. The full barycentric correction ($\lambda = 1$) with respect to $D_1$ is $Z_{1}=Z-aD_1$. Define
\[
\Delta \Ucal_{2}(1)= \mathcal E_{1,2}(1) =\Ucal_2(Z_{1})-\Ucal_2(Z),
\qquad s=\sqrt{(1-a^2)(1-\rho^2)}.
\]
Then:

(1) \emph{Sharp bounds on the externality.} The identified set for $\Delta \Ucal_{2}(1)$ is exactly
\begin{equation}\label{eq:hidden-sharp-fr}
{\ -a^2\rho^2-2|a\rho|s\ \leq\ \Delta \Ucal_{2}(1)
\ \leq\ -a^2\rho^2+2|a\rho|s.\ }
\end{equation}

(2) \emph{Central value under conditional independence.} The arithmetic midpoint of the bounds in \eqref{eq:hidden-sharp-fr} is attained under the assumption $D_2\perp Z\mid D_1$, in which case
\begin{equation}\label{eq:hidden-center-fr}
{\ \Delta \Ucal_{2}(1)=-a^2\rho^2.\ }
\end{equation}

(3) \emph{Residual disparity.} After full correction for $D_1$, the identified set for the disparity associated with the unobserved group is exactly
\begin{equation}\label{eq:hidden-resid-fr}
{\ 0\ \leq\ \Ucal_2(Z_{1})\ \leq\ (1-a^2)(1-\rho^2).\ }
\end{equation}
All bounds are \emph{attainable} among Gaussian laws, including possibly degenerate Gaussian laws, compatible with the observed correlations $(a,r)$. The proof is given in Appendix \ref{app:subsec:proof-prop-3-1}.
\end{proposition}
We define the unknown correlation parameter $\eta=\operatorname{Corr}(D_2,Z\mid D_1)$. Using this, we can rewrite $\Delta \Ucal_{2}(1)$ as
\begin{equation}\label{eq:hidden-delta-eta-fr-main}
\Delta \Ucal_{2}(1)
=(b-a\rho)^2-b^2
=-a^2\rho^2-2a\rho\eta s.
\end{equation}

\paragraph{Interpreting the central curve.}
Equation \eqref{eq:hidden-delta-eta-fr-main} separates the term $-a^2\rho^2$, which is determined by the known correlations, from the term involving the unknown partial correlation $\eta$. The central value ($\eta=0$) corresponds to $D_2 \perp Z \mid D_1$. It is neither an estimation nor, without an additional assumption, a probabilistic average. It can be interpreted as an expectation only after specifying a model for uncertainty about $\eta$ under which $\E[\eta]=0$.

\begin{proposition}[A certificate of non-deterioration]
\label{prop:hidden-certificate-fr}
Under the assumption of Proposition \ref{prop:hidden-sharp-fr},
the correction targeting $D_1$ does not increase the disparity associated with $D_2$ for \emph{any} Gaussian law compatible with $(a,\rho)$ if and only if
\[
 a\rho=0\quad\text{or}\quad
 4(1-a^2)(1-\rho^2)\leq a^2\rho^2.
\]
If $a\rho\neq0$ and the second inequality fails, some compatible laws yield an increase while others yield a decrease. The proof is given in Appendix \ref{app:subsec:proof-prop-3-2}.
\end{proposition}

\paragraph{Two special cases.}
If $\rho=0$, then $D_1$ and $D_2$ are independent under the Gaussian model, and $\Delta \Ucal_{2}(1)=0$ regardless of the dependence between $D_2$ and $Z$. By contrast, if $\eta=0$, then $D_2\perp Z\mid D_1$ and $\Delta \Ucal_{2}(1)=-a^2\rho^2$. Finally, \emph{no initial disparity} with respect to $D_2$ corresponds to $b=0$. Whenever this is compatible with $(a,\rho)$, it yields $\Delta \Ucal_{2}(1)=a^2\rho^2>0$ provided $a\rho\neq0$. Thus, a correction can create disparity that was initially absent.

\begin{figure}[t]
\centering
\includegraphics[width=0.8\linewidth]{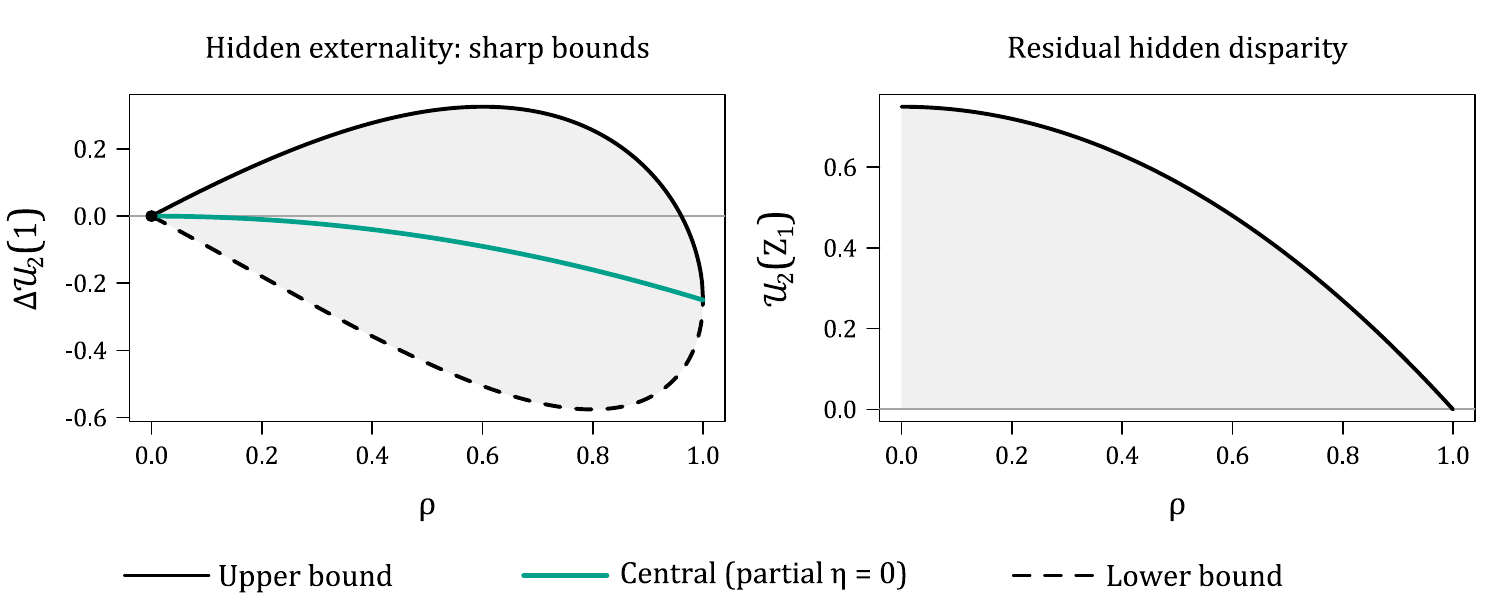}
\caption{\textbf{Partial identification of a hidden externality.}
We fix $a=\operatorname{Corr}(D_1,Z)=0.5$ and vary the correlation $\rho=\operatorname{Corr}(D_1,D_2) \in [0,1]$ along the horizontal axis. The left panel shows the sharp bounds on the change in disparity associated with $D_2$. The central blue curve is their arithmetic midpoint and corresponds to conditional independence ($\eta=0$); the exterme curves correspond to $\eta=\pm1$. The right panel shows the sharp identified interval for residual disparity after full correction of $D_1$, $[0,(1-a^2)(1-\rho^2)]$.}
\label{fig:hidden-bounds-fr}
\end{figure}

Figure~\ref{fig:hidden-bounds-fr} illustrates the results of Proposition~\ref{prop:hidden-sharp-fr} on the partial identification of the externality induced by correcting for $D_1$ when the sensitive attribute $D_2$ is unobserved. The left panel shows the sharp bounds on the change in disparity associated with $D_2$, $\Delta \Ucal_{2}(1)$, while the right panel displays the identified interval for the residual disparity, $\Ucal_2(Z_1)$, for different values of the correlation $\rho\in[0,1]$ between $D_1$ and $D_2$. As the correlation between the two sensitive attributes increases, the identified interval for $\Delta \Ucal_{2}(1)$ initially widens (for $\rho\in[0,0.8]$) and then shrinks (for $\rho>0.8$). Under conditional independence ($\eta=0$), the externality corresponds to the arithmetic midpoint of the sharp bounds and is nonpositive. For most values of $\rho$, the bounds straddle zero, indicating that the direction of the externality cannot always be determined from the observed correlations alone. However, as $\rho\to 1$, $\Delta \Ucal_{2}(1)$ is necessarily negative in this example, indicating that correcting for $D_1$ also reduces disparity with respect to $D_2$. In the right panel, the identified interval for the residual disparity after full correction ($\lambda=1$) narrows as the correlation between the sensitive attributes approaches one, eventually collapsing to zero.

\subsection{Generalization to Categorical Sensitive Attributes}
\label{subsec:sharp-hidden-categorical}

We now return to a more general setting in which both sensitive attributes are categorical, with $D_1\in\{1,\dots,K\}$ and $D_2\in\{0,1\}$. The joint law of the observed variables $(D_1,Z)$ is available, where $Z$ is a real-valued, square-integrable score. An external source also provides the proportions $q_i=\mathbb P(D_2=1\mid D_1=i)$ for the unobserved binary attribute $D_2$. Let $p_i=\mathbb P(D_1=i)$ and $q=\sum_{i=1}^K p_iq_i\in(0,1)$. Write $F_i=\mathcal L(Z\mid D_1=i)$, and let $T_i$ denote the monotone transport map from $F_i$ to their Wasserstein barycenter, assumed to be well defined. The corrected score is $Z'=T_{D_1}(Z)$.


\begin{proposition}[Exact partial identification]
\label{prop:sharp-hidden-categorical}
Let $\mathcal H$ bet the set of families of positive sub-probability measures $H=(H_i)_{i=1}^K$ satisfying
\[
0\le H_i\le F_i,\qquad H_i(\mathbb R)=q_i.
\]
For each $H\in\mathcal H$, define
\[
\begin{aligned}
 P_{1,H}&=\frac1q\sum_i p_iH_i,&
 P_{0,H}&=\frac1{1-q}\sum_i p_i(F_i-H_i),\\
 P'_{1,H}&=\frac1q\sum_i p_i(T_i)_\#H_i,&
 P'_{0,H}&=\frac1{1-q}\sum_i p_i(T_i)_\#(F_i-H_i).
\end{aligned}
\]
The identified set for the externality $\Delta \mathcal U_2=\mathcal U_2(Z_1)-\mathcal  U_2(Z)$ is exactly the interval
\begin{equation}\label{eq:sharp-hidden-population}
\left[\min_{H\in\mathcal H}\Psi(H),\;
             \max_{H\in\mathcal H}\Psi(H)\right],
\end{equation}
where $\Psi(H)=q(1-q)\left\{
W_2^2(P'_{1,H},P'_{0,H})-
W_2^2(P_{1,H},P_{0,H})\right\}$. Endpoints are attainable, and all intermediate values are compatible with the observed information. The proof is given in Appendix \ref{app:subsec:proof-prop-3-3}. 
\end{proposition}

\paragraph{Explicit distributional envelopes.} 
For every $z\in\mathbb R$, the constraints on $H_i$ imply
\[
\max\{0,F_i(z)-(1-q_i)\}
\le h_i(z)= H_i\left((\! -\infty,z]\right)\le\min\{F_i(z),q_i\},
\]
where $F_i(z)$ denotes the CDF associated with the measure $F_i$. It follows that the CDF of $P_{1,H}$ satisfies
\begin{equation}\label{eq:sharp-cdf-bounds}
\quad
\frac{\sum_i p_i\bigl[F_i(z)-(1-q_i)\bigr]_+}{q}
\le F_{1,H}(z)\le
\frac{\sum_i p_i\min\{F_i(z),q_i\}}q.
\end{equation}
Both envelopes can be attained simultaneously for every $z$ by assigning $D_2=1$, within each $\{D_1=i\}$ group, respectively to the $q_i$ highest or the $q_i$ lowest scores. Analogous envelopes hold for $P_{0,H}$ and for the corrected score distributions, replacing $F_i$ with $(T_i)_\#F_i$. These envelopes can be used to construct conservative numerical bounds on Wasserstein distances, although they do not generally yield a closed-form expression for \eqref{eq:sharp-hidden-population}. The empirical derivation of these bounds is presented in Appendix~\ref{app:subsec:empirical-bounds}, along with an illustrative example.

\section{Numerical Experiments on Real Data}\label{sec:results}

We investigate cross-attribute externalities across nine classification and regression datasets, following the application scenario in Section~\ref{sec:hidden-attribute-bounds}. Practitioners are assumed to have access to one or two categorical sensitive attributes, which they mitigate using Wasserstein barycentric correction. We assess the impact on a hidden attribute using the marginal and joint Wasserstein disparities defined in Section~\ref{subsec:notations}. Decision scores $Z$ are obtained from ML models trained without the hidden attribute, which is retained only for retrospective evaluation. In this setting, theoretical bounds under a jointly Gaussian model are illustrated in Appendix~\ref{app:subsec:joint-gaussian-ex-H}. We do not compute the sharp empirical bounds introduced in Appendix~\ref{app:subsec:empirical-bounds}, as this requires optimization over a potentially combinatorial number of feasible hidden-attribute assignments, although without retraining the ML models since the original and corrected scores remain fixed. Instead, we directly evaluate externalities using the withheld hidden attribute. Wasserstein-based evaluation and correction are implemented using the Python package \textbf{EquiPy} \cite{machado2025equipysequentialfairnessusing}.

\subsection{Binary Classification}

Following \cite{chen2024multiple}, we consider five datasets covering prediction tasks in sensitive decision-making domains (income, healthcare, payment default, and judicial): Adult, COMPAS, Default, Mep15, and Mep16. 
For each combination of dataset and sensitive attributes available to the practitioner, we train four ML models: Logistic Regression (LR), Random Forest (RF), Support Vector Machine (SVM), and Deep Neural Network (DNN). We first consider tasks in which the practitioner mitigates one sensitive attribute while another remains hidden, focusing on sex and race. For Mep15 and Mep16, we additionally consider age, defined as being older than 65, as a sensitive attribute. Results for this setting, with race as the hidden attribute, are reported in Appendix~\ref{app:subsec:binary-classification}. To mitigate unfairness when intervening on one attribute, we compare \textbf{EquiPy}, which specifically targets Wasserstein disparities, with two additional methods in Appendix~\ref{app:subsec:binary-classification}: a reweighting pre-processing method (\textbf{RW} \cite{kamiran2012data}) and a rank-preserving approach (\textbf{DIR} \cite{feldman2015certifying}).

\begin{table}[t]
\centering
\caption{\textbf{Changes in hidden-attribute unfairness, $\Delta \Ucal_h$, under Wasserstein barycenter mitigation (\textbf{EquiPy}).} For each dataset and mitigated--hidden attribute pair, reports include the mean change (standard deviation in parentheses), minimum, maximum, and \% of runs with $\Delta \Ucal_h>0$ (indicating a deterioration in fairness), across 30 random seeds and four classification models (LR, RF, SVM, and DNN). $\rho_{1h}$ is the correlation between the mitigated and hidden attributes.}
\label{tab:delta-unf-summary-two-attr-binary}
\small
\setlength{\tabcolsep}{5pt}
\begin{tabular}{ll l r r r r r r}
\toprule
\textbf{Dataset} &
\textbf{Mitigated} &
\textbf{Hidden} &
$\boldsymbol{\rho_{1h}}$ & 
\textbf{Mean $\Delta \Ucal_h$ (sd)} &
\textbf{Min $\Delta \Ucal_h$} &
\textbf{Max $\Delta \Ucal_h$} &
\textbf{\% $\Delta \Ucal_h$ > 0} \\
\midrule
Adult & Race & Sex & 0.101 & -0.019 (0.013) & -0.060 & 0.005 & 6.7 \\
& Sex & Race & 0.101 & -0.059 (0.020) & -0.111 & 0.003 & 0.8 \\
\midrule
Compas & Race & Sex & 0.066 & -0.015 (0.011) & -0.051 & 0.018 & 5.0 \\
& Sex & Race & 0.066 & -0.011 (0.007) & -0.033 & 0.004 & 5.8 \\
\midrule
Default & Race & Sex & -0.068 & \bf0.002 (0.007) & -0.012 & 0.037 & 56.7 \\
& Sex & Race & -0.068 & \bf0.002 (0.007) & -0.025 & 0.022 & 67.5 \\
\midrule
Mep15& Race & Sex & -0.017 & \bf0.003 (0.008) & -0.021 & 0.024 & 65.0 \\
& Sex & Race & -0.017 & \bf0.006 (0.008) & -0.011 & 0.031 & 75.8 \\
\midrule
Mep16 & Race & Sex & -0.018 & -0.001 (0.007) & -0.020 & 0.015 & 45.0 \\
& Sex & Race & -0.018 & -0.001 (0.010) & -0.039 & 0.021 & 55.8 \\
\bottomrule
\end{tabular}
\end{table}

Table~\ref{tab:delta-unf-summary-two-attr-binary} summarizes changes in hidden-attribute unfairness, $\Delta \mathcal{U}_h$, when mitigating sex or race using \textbf{EquiPy}. Results vary across datasets: for Adult and COMPAS, the mean change is negative in both mitigation directions, indicating reduced hidden-attribute unfairness; in contrast, Default and Mep15 exhibit positive near-zero mean changes, with a notable proportion of runs showing $\Delta \mathcal{U}_h>0$. For Mep16, despite having slightly negative mean changes in both directions, hidden-attribute unfairness increases in 56\% of runs when mitigating sex. 
Since $|\rho_{1,h}|$ is greater for Mep16 than for Mep15, pairwise correlation alone is insufficient to determine the direction of the externality. Cross-attribute externalities depend not only on the statistical dependence between sensitive attributes but also on how predictive scores vary across their joint profiles.

\begin{figure}[t]
\centering\includegraphics[width=.7\linewidth]{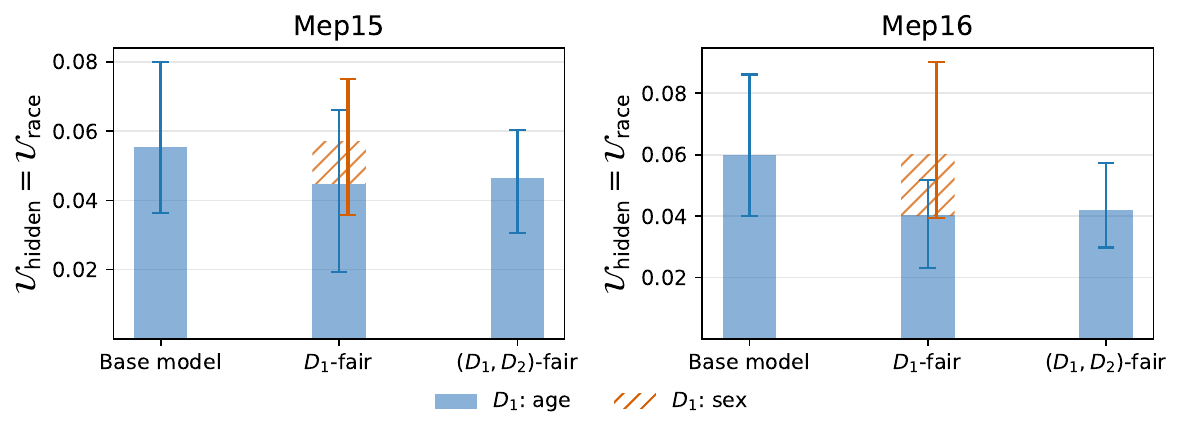}
\caption{\textbf{Intermediate and joint mitigation effects on hidden-attribute unfairness with \textbf{EquiPy}.} Mitigating both age and sex, we report race-related unfairness for the baseline model, after age or sex mitigation, and after joint age--sex mitigation, for Mep15 (left panel) and Mep16 (right panel). Individual mitigation steps can reduce hidden-attribute unfairness, whereas the subsequent joint correction can increase it (below the baseline level). Error bars and markers indicate variability across seeds and models.}
\label{fig:binary-classif-lr-u3}
\end{figure}

In the setting with two observed and one hidden attribute, Figure~\ref{fig:binary-classif-lr-u3} shows the hidden-attribute unfairness at each stage of the mitigation sequence for age and sex, using the LR model. Both Mep15 and Mep16 exhibit similar patterns: mitigating the first attribute reduces race-related unfairness, whereas mitigating the second leads to an increase. This illustrates the non-monotonic effects of sequentially mitigating multiple sensitive attributes on hidden-attribute fairness.
\subsection{Regression}

We consider four real-world datasets in the evaluation of fairness for regression tasks \citep{almajed2024consistencyfairnessmeasurementmethods}. They cover different application domains, including educational outcomes \citep{Wightman1998LSACNL}, community-level crime rates \citep{Redmond2009}, medical insurance costs \citep{lantz2015machine}, and individual income \citep{ding2021retiring}. 
Table \ref{tab:delta_uh_summary_sens1} summarizes the changes in hidden-attribute unfairness, $\Delta \Ucal_h$, when a single observed attribute is mitigated while another remains hidden. For each mitigated-hidden attribute pair, Table \ref{tab:delta_uh_summary_sens1} presents one mitigation direction, while the corresponding reverse directions are reported in Table \ref{tab:delta_uh_summary_sens2} in Appendix. Overall, the externalities depend on the dataset and the specific mitigated-hidden attribute pair. The most consistent reductions are observed in the Communities and Crime dataset when poverty and race are considered as the mitigated and hidden attributes, respectively, or vice versa. In both directions, the mean change is negative ($-0.164$ and $-0.181$, respectively), and $\Delta \Ucal_h$ is negative in all evaluated runs. However, mitigating unfairness with respect to one observed attribute does not necessarily improve fairness with respect to a hidden attribute across all runs. In Communities and Crime, mitigating age when poverty is hidden produces a mean change of $-0.019$, while 22.2\% of runs exhibit an increase; in the reverse direction, the mean change is positive ($0.002$), with an increase in 54.4\% of runs. In ACSIncome, when age and sex are considered, the mean changes are positive in both mitigation directions ($0.029$ and $0.006$), and increases occur in 66.7\% and 60.0\% of runs, respectively. Figure \ref{fig:z_lambda_unfairness_comparison} in Appendix illustrates that improvements in the targeted and joint criteria need not translate into a monotonic reduction in hidden-attribute unfairness.

\begin{table}[t]
\centering
\caption{\textbf{Changes in hidden-attribute unfairness, $\Delta \Ucal_h$, under Wasserstein barycenter mitigation (\textbf{EquiPy})}. For each dataset and mitigated--hidden attribute pair, $\Delta \Ucal_h$ is summarized across 30 random seeds and the regression models (GLM, RF, and NN) considered. The table reports the mean change (standard deviation in parentheses), minimum, maximum, and percentage of runs with $\Delta \Ucal_h>0$, and $\rho_{1h}$ denotes the correlation between the mitigated and hidden attributes. Positive values indicate an increase in hidden-attribute unfairness. The reverse mitigation directions are reported in Table \ref{tab:delta_uh_summary_sens2} in Appendix.}
\label{tab:delta_uh_summary_sens1}
\small
\setlength{\tabcolsep}{5pt}
\begin{tabular}{ll l r r r r r r}
\toprule
\textbf{Dataset} &
\textbf{Mitigated} &
\textbf{Hidden} &
$\boldsymbol{\rho_{1h}}$ & 
\textbf{Mean $\Delta \Ucal_h$ (sd)} &
\textbf{Min $\Delta \Ucal_h$} &
\textbf{Max $\Delta \Ucal_h$} &
\textbf{\% $\Delta \Ucal_h$ > 0} \\
\midrule
Law School & Race & Sex & 0.077 & -0.035 (0.059)& -0.184 & 0.133 & 23.3 \\
\midrule
Insurance & Sex & Age & 0.003 & -0.138 (0.240) & -0.802 & 0.396 & 27.8 \\
\midrule
Communities and Crime & Race & Age & 0.085 & \bf0.002 (0.043) & -0.089 & 0.169 & 50.0 \\%
& Age & Poverty & 0.138 & -0.019 (0.034) & -0.105 & 0.149 & 22.2 \\
& Poverty & Race & -0.336 & -0.164 (0.026) & -0.235 & -0.118 & 0.0 \\
\midrule
ACSIncome (PA) & Race & Sex & 0.022 & -0.027 (0.072) & -0.358 & 0.121 & 34.4\\
& Age & Sex & 0.012 & \bf0.029 (0.092) & -0.239 & 0.353 & 66.7  \\
& Race & Age & 0.090 & -0.059 (0.079) & -0.255 & 0.127 & 22.2 \\
\bottomrule
\end{tabular}
\end{table}

\begin{figure}[htbp!]
    \centering
    \includegraphics[width=0.4\linewidth]{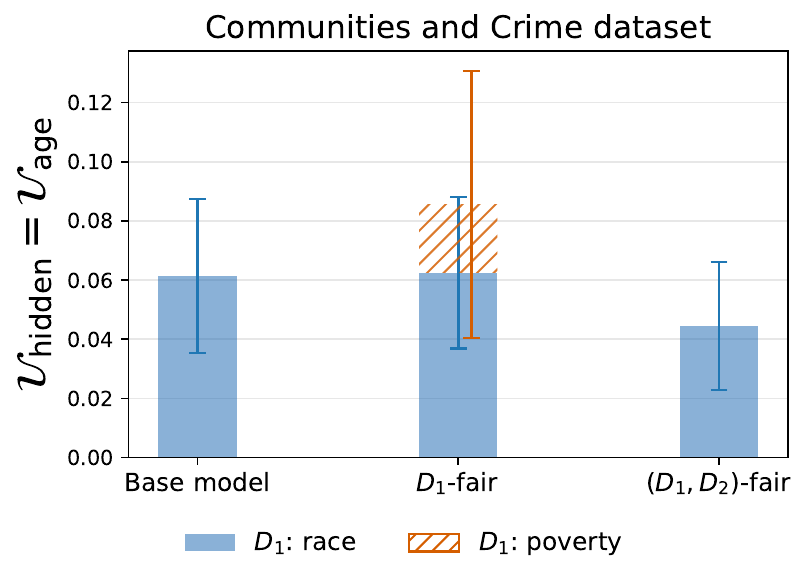}
    \includegraphics[width=0.4\linewidth]{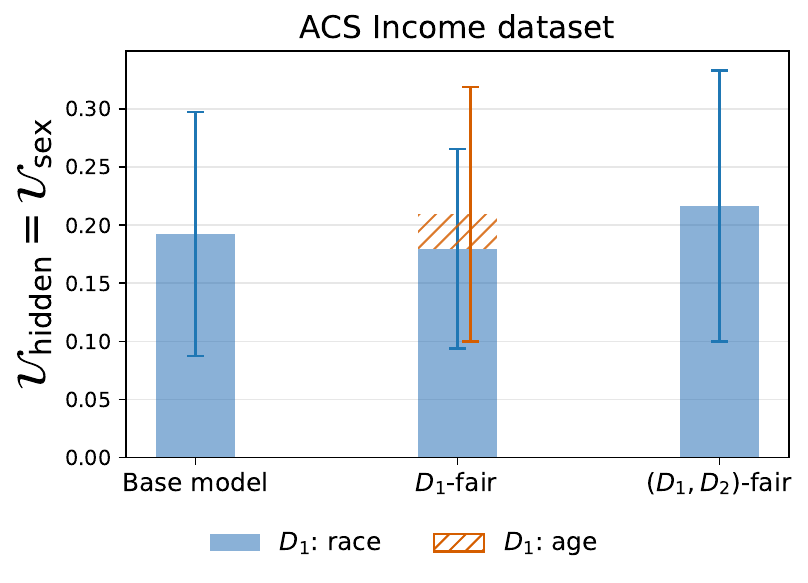}
    \caption{\textbf{Intermediate and joint mitigation effects on hidden-attribute unfairness.} The left panel reports age-related unfairness for the baseline model, after race or poverty mitigation, and after joint race--poverty mitigation in the Communities and Crime dataset. The right panel reports sex-related unfairness for the baseline model, after race or age mitigation, and after joint race-age mitigation in the ACS Income dataset. The plots illustrate that an individual mitigation step can increase unfairness associated with a hidden attribute, whereas the subsequent joint correction can reduce unfairness below its baseline level. Error bars and markers indicate the variability or range across the evaluated results (across seeds and models).}
    \label{fig:crime_sequential}
\end{figure}

Table~\ref{tab:delta_uh_summary3} summarizes $\Delta \Ucal_h$ when two attributes are mitigated while another one is hidden. 
In ACSIncome, when race and age are mitigated, unfairness with respect to hidden sex increases on average ($\Delta \Ucal_h=0.024$), and in a majority of the evaluated configurations ($\Delta \Ucal_h>0$ in 64.4\% of runs). Figure~\ref{fig:crime_sequential} highlights the mitigation sequence when working with three sensitive attributes, with the distinction between the externalities of individual mitigation steps and the cumulative effects of joint mitigation. In the Communities and Crime dataset (left panel), the final jointly corrected model exhibits lower unfairness than the baseline, although the intermediate result indicates that the effect of poverty increases the hidden-attribute unfairness. For ACSIncome (right panel), the intermediate age-correction increases unfairness with respect to hidden sex, and this increase is not fully offset by the joint correction. 
Figure \ref{fig:model_dependent} in Appendix further illustrates that the fairness externalities of mitigation are model-dependent: the same mitigation procedure can produce different intermediate and cumulative effects depending on the predictive model to which it is applied.

\begin{table}[t]
\centering
\caption{\textbf{Changes in hidden-attribute unfairness, $\Delta \Ucal_h$, under Wasserstein barycenter mitigation (\textbf{EquiPy})}. For each dataset, the table considers pairs of mitigated attributes $(D_1, D_2)$ and a hidden attribute $h$. Results are summarized across 30 random seeds and the regression models (GLM, RF, and NN). The table reports the mean change (standard deviation in parentheses), minimum, maximum, and percentage of runs with $\Delta \Ucal_h>0$, and $\rho_{1h}$, $\rho_{2h}$ denote the correlation between the hidden attribute $h$ and the first and second mitigated attribute, respectively. Positive values of $\Delta \Ucal_h$ indicate an increase in hidden-attribute unfairness.}
\label{tab:delta_uh_summary3}
\small
\setlength{\tabcolsep}{5pt}
\begin{tabular}{ll l r r r r r r}
\toprule
\textbf{Dataset} &
\textbf{Mitigated} &
\textbf{Hidden} &
$\boldsymbol{\rho_{1h}}$ & 
$\boldsymbol{\rho_{2h}}$ &
\textbf{Mean $\Delta \Ucal_h$} &
\textbf{Min $\Delta \Ucal_h$} &
\textbf{Max $\Delta \Ucal_h$} &
\textbf{\% $\Delta \Ucal_h$ > 0} \\
\midrule
Communities and Crime & Race/Age & Poverty & -0.3358 & 0.1383 & -0.193 (0.054) & -0.316 & -0.057 & 0.0 \\
& Race/Poverty & Age & 0.0853 & 0.1383 & -0.014 (0.036) & -0.097 & 0.089 & 28.9 \\
& Age/Poverty & Race & 0.0853 & -0.3358 & -0.175 (0.025) & -0.240 & -0.115 & 0.0 \\
\midrule
ACSIncome (PA) & Race/Age & Sex & 0.0219 & 0.0123 & \bf0.024 (0.112) & -0.260 & 0.336 & 64.4 \\
& Sex/Age & Race & 0.0219 & 0.0896 & -0.547 (0.368) & -1.340 & 0.120 & 3.3 \\
& Sex/Race & Age & 0.0123 & 0.0896 & -0.049 (0.082) & -0.275 & 0.2325 & 22.75 \\
\bottomrule
\end{tabular}
\end{table}

\section{Conclusion}

To conclude, this work highlights the importance of accounting for cross-attribute externalities when designing and evaluating fairness interventions. Our theoretical and empirical results show that mitigating disparities associated with one sensitive attribute can exacerbate unfairness with respect to other attributes, including those unavailable to the practitioner. These effects can be complex and non-monotonic, depending on the underlying data distribution, the type of model used, and the sequence of interventions when more than one sensitive attribute is available. Our findings therefore underscore the limitations of evaluating fairness solely with respect to the targeted attribute and motivate a more comprehensive assessment of marginal and joint disparities across multiple sensitive attributes.




\section*{Acknowledgments}

Agathe Fernandes Machado acknowledges that the project leading to this publication has received funding from OBVIA. Arthur Charpentier acknowledges funding from the SCOR Foundation for Science and the National
Sciences and Engineering Research Council (NSERC) for funding (RGPIN-2019-07077).

\nocite{agueh2011barycenters,chzhen2020fair,gordaliza2019obtaining,hu2024sequential,fernandesmachado2026aistats}
\bibliographystyle{plainnat}
\bibliography{biblio-facct}

\appendix

\section{Proofs}

\subsection{Proof of Proposition \ref{prop:target}}\label{app:subsec:proof-prop-2-1}
\begin{proof} The corrected quantile in group $a$ is $(1-\lambda)Q_{1,a}+\lambda\bar Q_1$; its  barycenter remains $\bar Q_1$. It follows from \eqref{eq:Um}.
\end{proof}

\subsection{Proof of Proposition \ref{prop:jointbound}}\label{app:subsec:proof-prop-2-2}
\begin{proof} For any reference law $\nu$, the convexity of $P\mapsto\W^2(P,\nu)$ gives, by grouping profiles according to $D_1=a$, $\sum_{a,b}p_{ab}\W^2(\nu_{ab},\nu)\ge\sum_a p_a\W^2(\nu_a,\nu)$. We then minimize with regards to $\nu$; the other marginals are treated similarly. This result covers the audits \emph{levels}, not the signs of their variations after intervention.
\end{proof}


\subsection{Proof of Proposition \ref{prop:gauss}}\label{app:subsec:proof-prop-2-3}
\begin{proof}
For any square-integrable random variable $V$ and real numbers $a,b$,
\[
\W^2\!\left(\mathcal L(V+a),\mathcal L(V+b)\right)=(a-b)^2.
\]
Indeed, coupling the two translations through the same $V$ gives the upper bound, while the difference between their means gives the matching lower bound. By \eqref{eq:gauss} and Gaussian conditioning of $(D_1,D_2)$, the distributions of $Z_\lambda$ conditional on either $D_j$ or on $(D_1,D_2)$ form location families with a common shape. Their barycentric indices consequently equal the variances of their conditional means.

Specifically,
\[
\mathbb E[Z_\lambda\mid D_1]=(1-\lambda)c_1D_1,
\qquad
\mathbb E[Z_\lambda\mid D_2]=(c_2-\lambda\rho c_1)D_2,
\]
which yields the first two expressions in \eqref{eq:paths}. Also,
\[
\mathbb E[Z_\lambda\mid D_1,D_2]
=(\beta_1-\lambda c_1)D_1+\beta_2D_2.
\]
Taking its variance and using $c_1=\beta_1+\rho\beta_2$ yields
\[
\Ucalj(Z_\lambda)
=\beta_1^2+\beta_2^2+2\rho\beta_1\beta_2
-\lambda(2-\lambda)c_1^2.
\]
Finally, subtracting $\Ucal_2(Z)=c_2^2$ from the second expression in \eqref{eq:paths} gives \eqref{eq:gaussian_externality}.
\end{proof}

\subsection{Proof of Proposition~\ref{prop:hidden-sharp-fr}}
\label{app:subsec:proof-prop-3-1}

\begin{proof}
The correlation matrix
\[
\Sigma=\begin{pmatrix}1&\rho&a\\\rho&1&b\\a&b&1\end{pmatrix}
\]
must be positive semi-definite. Since $|a|,|\rho|<1$, this is equivalent to
\[
\det(\Sigma)=(1-a^2)(1-\rho^2)-(b-a\rho)^2\geq0,
\]
Consequently, $b\in[a\rho-s,a\rho+s]$. Equivalently, there exists $\eta\in[-1,1]$ such that
\begin{equation}\label{eq:hidden-partial-fr}
b=a\rho+\eta s,
\qquad
\eta=\operatorname{Corr}(D_2,Z\mid D_1),
\end{equation}
where $\eta$ is the Gaussian partial correlation.

The conditional distributions of $Z$ given $D_2$ are Gaussian translations with common variance. Hence the Wasserstein index equals the variance of the conditional mean:
\[
\Ucal_2(Z)=b^2,
\qquad
\Ucal_2(Z_{1})=(b-a\rho)^2.
\]
Subtracting and applying \eqref{eq:hidden-partial-fr} gives,
\begin{equation}\label{eq:hidden-delta-eta-fr}
\Delta \Ucal_2(1)
=(b-a\rho)^2-b^2
=-a^2\rho^2-2a\rho\eta s.
\end{equation}
The bounds in \eqref{eq:hidden-sharp-fr} follow from $\eta\in[-1,1]$.
Under joint Gaussianity, $D_2\perp Z\mid D_1$ implies $\eta=0$, which gives \eqref{eq:hidden-center-fr}. Finally,
\[
\Ucal_2(Z_1)=(b-a\rho)^2=\eta^2s^2,
\]
which establishes \eqref{eq:hidden-resid-fr}.

It remains to show sharpness. Let $X,E,W$ be independent standard Gaussian variables and, for any $\eta\in[-1,1]$, define
\[
\begin{aligned}
D_1&=X,\\
D_2&=\rho X+\sqrt{1-\rho^2}\,E,\\
Z&=aX+\sqrt{1-a^2}
   \bigl(\eta E+\sqrt{1-\eta^2}\,W\bigr).
\end{aligned}
\]
This construction reproduces the specified correlations $a,\rho$, while $\operatorname{Corr}(D_2,Z)=a\rho+\eta s$. Every $\eta\in[-1,1]$ is therefore compatible with the available information, and every point in the stated intervals is attained.
\end{proof}

\subsection{Proof of Proposition \ref{prop:hidden-certificate-fr}}\label{app:subsec:proof-prop-3-2}

\begin{proof}
Uniform non-deterioration 
is equivalent to requiring that the upper bound in \eqref{eq:hidden-sharp-fr} be non-positive: $-a^2\rho^2+2|a\rho|s\leq0$. If $a\rho=0$, the externality is identically zero. Otherwise, the condition is equivalent to $2s\leq|a\rho|$, which gives the stated criterion after squaring. If the criterion fails, the upper bound is positive and the lower bound is strictly negative. Both signs are attainable by the construction in the proof of Propostion~\ref{prop:hidden-sharp-fr}.
\end{proof}

\subsection{Proof of Proposition~\ref{prop:sharp-hidden-categorical}} 
\label{app:subsec:proof-prop-3-3}
\begin{proof}
Every compatible joint law induces sub-measures $H_i(A)=\mathbb P(Z\in A,D_2=1\mid D_1=i)$, which satisfy the stated constraints: $H_i(\mathbb R)=q_i, 0\le H_i\le F_i$. Conversely, for every $H_i\le F_i$, the Radon--Nikodym derivative $w_i=dH_i/dF_i\in[0,1]$ defines an admissible assignment mechanism $\mathbb P(D_2=1\mid D_1=i,Z=z)=w_i(z)$. Thus, the four conditional distributions are exactly those specified in the proposition. For a binary attribute, the one-dimensional barycentric identity gives $\Ucal_2=q(1-q)W_2^2(P_1,P_0)$, which establishes \eqref{eq:sharp-hidden-population}.

The set $\mathcal H$ is convex and compact unver weak convergence. The domination $H_i\le F_i$, together with the finite second moments of $F_i$ and $(T_i)_\#F_i$, ensure the required continuity in $W_2$. Hence $\Psi$ is continuous and attain its miminum and maximum. Finally, the continuous image of a connected set is connected; since $\mathcal H$ is convex and therefore connected, its image under $\Psi$ is an interval. 
\end{proof}

\subsection{Proof of Proposition \ref{prop:sharp-hidden-sample}} \label{app:subsec:proof-prop-3-4}
\begin{proof}
For two empirical distributions of equal size $n/2$, the monotone coupling pairs their ordered statistics, giving
\[
W_2^2=\frac2n\sum_{k=1}^{n/2}(z_{(k),1}-z_{(k),0})^2.
\]
Because the two hidden groups have equal probability ($q = 1/2$), $\mathcal U_2=\frac14W_2^2$. Since the hidden assignment $D_2$ is the only unknown quantity, enumerating $\mathcal A$ yields exactly all admissible values.
\end{proof}

\section{An explicitly solvable Gaussian model}
\subsection{Three Sensitive Attributes}\label{app:subsec:three_sensitive}

Consider a standard Gaussian random vector $(D_1,D_2,D_3)$, with unit marginal variances
and correlation matrix
$\Sigma_D = (\rho_{ij})
\succeq 0$,
where $\rho_{ij}=\mathrm{Corr}(D_i,D_j)\in(-1,1)$
for $i\neq j$. Let the additive score be
\begin{equation}\label{eq:gauss3}
Z=\beta_1D_1+\beta_2D_2+\beta_3D_3+R_{\boldsymbol{D}},\qquad
R_{\boldsymbol{D}}\sim\mathcal N(0,\sigma^2),\quad R_{\boldsymbol{D}}\perp(D_1,D_2,D_3).
\end{equation}
Define $c_j=\mathrm{Cov}(Z,D_j)$, such that $c_1=\beta_1+\rho_{12}\beta_2+\rho_{13}\beta_3$, $c_2=\rho_{12}\beta_1+\beta_2+\rho_{23}\beta_3$, and $c_3=\rho_{13}\beta_1+\rho_{23}\beta_2+\beta_3$. Since the conditional distributions of $Z$ given $D_1$ are Gaussian distributions with a common variance and conditional means $c_1D_1$, the barycentric repair targeting $D_1$ is exactly
\begin{equation}\label{eq:gaussrepair3}
Z_\lambda=Z-\lambda c_1D_1 \qquad \lambda\in[0,1].
\end{equation}
At $\lambda=0$, the original score is recovered; at $\lambda=1$, the conditional mean of the score given $D_1$ is completly removed.

\begin{proposition}[Four explicit trajectories]\label{prop:gauss3}
For the model above in \eqref{eq:gauss3}, the three marginal disparity indices and the joint disparity index evolve along the repair trajectory as
\begin{align}\label{eq:paths3}
\Ucal_1(Z_\lambda)&=(1-\lambda)^2c_1^2,\\
\Ucal_2(Z_\lambda)&=(c_2-\lambda\rho_{12} c_1)^2,\nonumber\\
\Ucal_3(Z_\lambda)&=(c_3-\lambda\rho_{13} c_1)^2,\nonumber\\
\Ucalj(Z_\lambda)&=\beta_1^2+\beta_2^2+\beta_3^2
+2\rho_{12}\beta_1\beta_2+2\rho_{13}\beta_1\beta_3+2\rho_{23}\beta_2\beta_3
 -\lambda(2-\lambda)c_1^2.\nonumber
\end{align}
The target disparity $\Ucal_1$ is eliminated at $\lambda=1$, and the joint disparity decreases monotonically throughout the repair trajectory. However, either or both of the non-target marginal disparities, $\Ucal_2$ and $\Ucal_3$, may increase, including immediately after repair begins.
\end{proposition}
\begin{proof} For two Gaussian distributions with the same variance, the squared Wasserstein distance $\W^2$ between them equals the squared difference between their means. Thus, each marginal disparity index is the variance of the corresponding conditional expectation $\E[Z_\lambda\mid D_1]$, $\E[Z_\lambda\mid D_2]$, $\E[Z_\lambda\mid D_3]$ et $\E[Z_\lambda\mid D_1,D_2,D_3]$.
Using the standard Gaussian conditional expectation identities, and since each $D_i$ has unit variance, taking the variance of these conditional expectations yields \eqref{eq:paths3}.
\end{proof}

At full correction ($\lambda=1$), the target disparity vanishes, while the two non-target marginal disparities become 
\begin{align} 
\mathcal{U}_2(Z_1) &=(c_2-\rho_{12}c_1)^2
=( \beta_2 (1- \rho_{12}^2 ) + \beta_3 (\rho_{23}  - \rho_{12}\rho_{13}) )^2,
\label{eq:full2}\\ \mathcal{U}_3(Z_1) &=(c_3-\rho_{13}c_1)^2
=(\beta_2(\rho_{23} - \rho_{13}\rho_{12}) +\beta_3 (1 -\rho_{13}^2) )^2 . \label{eq:full3} \end{align}
Their initial values are $\mathcal{U}_2(Z)=c_2^2$ and $\mathcal{U}_3(Z)=c_3^2$, respectively. Consequently, the externalities become positive when 
\begin{equation}\label{eq:region2}
|c_2-\rho_{12}c_1|>|c_2|, \qquad
|c_3-\rho_{13}c_1|>|c_3|,
\end{equation}
This phenomenon is a form of \emph{masking and unmasking}: before the intervention, the contributions associated with different correlated variables can partially cancel in a marginal audit. Removing the contribution associated with $D_1$ may destroy this compensation.


\paragraph{Analytical phase regions.}
Define $a_2=\rho_{12}c_1$ and $a_3=\rho_{13}c_1$.
For $j\in\{2,3\}$, the change in the corresponding marginal disparity is
\begin{equation}\label{eq:difference}
\mathcal{U}_j(Z_\lambda)-\mathcal{U}_j(Z)
= \lambda a_j(\lambda a_j-2c_j).
\end{equation}

For either non-target variable $D_j$, deterioration is \emph{immediate} if $a_jc_j<0$, or if $c_j=0$ and $a_j\neq 0$: the disparity increases for every $\lambda>0$.

Deterioration is \emph{delayed} if $a_jc_j>0$ and
\begin{equation}
0<\frac{2c_j}{a_j}<1.
\end{equation}
In this case, the disparity initially decreases, reaches its minimum at $\lambda_j^\star=c_j/a_j$, and subsequently increases, exceeding its initial value once $\lambda>2c_j/a_j$.

There is \emph{no strict deterioration} if $a_j=0$, in which case the disparity remains unchanged, or if $a_jc_j>0$ and $2c_j/a_j\geq 1$.
In the latter case, the disparity never exceeds its initial value on $[0,1]$, although it need not decrease monotonically. More precisely, it decreases and then increases when $1/2<c_j/a_j<1$, while remaining below its initial value; it decreases throughout $[0,1]$ when $c_j/a_j\geq 1$. At the boundary $2c_j/a_j=1$, it returns exactly to its initial value at $\lambda=1$.

\subsection{Example with a Hidden Sensitive Attribute}\label{app:subsec:joint-gaussian-ex-H}

Section~\ref{subsec:joint-gaussian-ex} examined how correcting a score with respect to one sensitive attribute can affect disparities associated with another. In this previous setting, both attributes were available in the underlying representation of the score. We now consider a different setting in which one sensitive attribute is not observed by the practitioner. The objective is to characterize the potential effects of a correction based on the observed attribute on disparities associated with the unobserved one. This setting is particularly relevant when sensitive attributes are unavailable because they are not collected, cannot legally be used, or are otherwise inaccessible to the practitioner.

Let $(D_1,D_2)$ denote two standard Gaussian sensitive attributes, with correlation $\rho\in(-1,1)$. The practitioner observes $D_1$ and a vector of explanatory variables $\boldsymbol{X} \in \mathbb{R}^d$, but does not observe $D_2$. The latter remains defined at the population level, allowing us to study its associated disparities even though they cannot be directly evaluated using the practitioner's observed data.

Consider the Gaussian score representation from Section~\ref{subsec:joint-gaussian-ex}
\begin{equation}\label{eq:gaussresidual}
Z=\beta_1D_1+\beta_2D_2+R_{\boldsymbol{D}},
\end{equation}
where
\[
\mathbb E[R_{\boldsymbol{D}}\mid D_1,D_2]=0,
\qquad
R_{\boldsymbol{D}}\indep(D_1,D_2).
\]
The independence condition follows from when the relevant variables are jointly Gaussian and $R_{\boldsymbol{D}}$ is the residual from the linear projection of $Z$ onto $(D_1,D_2)$. The residual $R_{\boldsymbol{D}}$ captures the remaining score heterogeneity after removing the linear associations with both sensitive attributes. It includes contributions from explanatory variables $\boldsymbol{X} \in \mathbb{R}^d$ that cannot be linearly explained by $(D_1,D_2)$, together with any independent noise. Using this, scores can be rewritten as
\begin{equation}\label{eq:gaussresidual-x}
Z=b_1D_1+b_2D_2+\boldsymbol{\theta}^\top \boldsymbol{X}.
\end{equation}

Now, in the current setting, since $D_2$ is unavailable, we cannot construct a correction directly from the decomposition in \eqref{eq:gaussresidual}. We instead express the score using the information available to the practitioner, namely $D_1$ and a vector of explanatory variables $\boldsymbol{X} \in \mathbb{R}^d$, 
\begin{equation}\label{eq:gaussresiduald1}
\tilde{Z}=\tilde{b}_1D_1+\tilde{\boldsymbol{\theta}}^\top \boldsymbol{X}.
\end{equation}
where we assume $(\boldsymbol{X},D_1,D_2)$ is jointly gaussian. Define the residualized explanatory variable
\[
 \boldsymbol{X}^\perp = \boldsymbol{X} - \mathbb E [\boldsymbol{X} \mid D_1] .
\]
Since $D_1$ is standardized and the variables $(\boldsymbol{X},D_1)$ are jointly Gaussian,
\[
    \mathbb E [\boldsymbol{X} \mid D_1] = \operatorname{Cov}(\boldsymbol X,D_1)D_1.
    \]
It follows that $\boldsymbol X^\perp\indep D_1$. Substituting this decomposition into the original score yields
\begin{equation}
    \tilde{Z}
=
\left(
\tilde{b}_1+
\tilde{\boldsymbol\theta}^\top
\operatorname{Cov}(\boldsymbol X,D_1)
\right)D_1
+
\tilde{\boldsymbol\theta}^\top\boldsymbol X^\perp.
\end{equation}
Define
\[
\alpha_1
=
\tilde{b}_1+
\tilde{\boldsymbol\theta}^\top
\operatorname{Cov}(\boldsymbol X,D_1).
\]
Suppose that $v_X=
\operatorname{Var}
\left(
\tilde{\boldsymbol\theta}^\top\boldsymbol X^\perp
\right)>0$. We introduce the standardized scalar proxy
\[
X^\star
=
\frac{
\tilde{\boldsymbol\theta}^\top\boldsymbol X^\perp
}{
\sqrt{v_X}
},
\]
and define $\alpha_2=\sqrt{v_X}$. The score then takes the form
\begin{equation}
\label{eq:gaussian_proxy_score}
\tilde{Z}=\alpha_1D_1+\alpha_2X^\star.
\end{equation}
If $v_X=0$, we set $\alpha_2=0$, and the proxy term vanishes. By construction, $X^\star$ is standard Gaussian and $X^\star\perp D_1$.
However, $X^\star$ need not be independent of the unobserved attribute $D_2$. We therefore introduce
\begin{equation}
\label{eq:proxy_correlations}
\operatorname{Corr}(D_1,D_2)=\rho,
\qquad
\operatorname{Corr}(D_1,X^\star)=0,
\qquad
\operatorname{Corr}(D_2,X^\star)=\tau,
\end{equation}
where $\tau$ quantifies the association between the proxy $X^\star$ and the unavailable sensitive attribute $D_2$. The corresponding correlation matrix of $(D_1,D_2,X^\star)$ is
\[
\Sigma=
\begin{pmatrix}
1 & \rho & 0\\
\rho & 1 & \tau\\
0 & \tau & 1
\end{pmatrix},
\]
which is positive semi-definite if and only if
\begin{equation}
\label{eq:proxy_feasibility}
\rho^2+\tau^2\leq 1 ,
\end{equation}
or, equivalently,
\begin{equation}
|\tau|\leq\sqrt{1-\rho^2}.
\end{equation}
By joint Gaussianity, the conditional expectation
is then given by
\begin{equation}
\label{eq:proxy_conditional_expectation}
\mathbb E[X^\star\mid D_1,D_2]
=
\frac{\tau}{1-\rho^2}(D_2-\rho D_1).
\end{equation}
It follows from \eqref{eq:gaussresidual} that
\begin{align}
\mathbb E[\tilde{Z}\mid D_1,D_2]
&=
\alpha_1D_1+
\alpha_2\mathbb E[X^\star\mid D_1,D_2]
\nonumber\\
&=
\left(
\alpha_1-
\frac{\alpha_2\rho\tau}{1-\rho^2}
\right)D_1
+
\frac{\alpha_2\tau}{1-\rho^2}D_2.
\end{align}
Thus, the representation in \eqref{eq:gaussresidual} can be recovered with
\begin{equation}
\label{eq:beta_alpha_relation}
\begin{aligned}
&\beta_1=
\alpha_1-
\frac{\alpha_2\rho\tau}{1-\rho^2}, 
&\beta_2 =
\frac{\alpha_2\tau}{1-\rho^2}.
\end{aligned}
\end{equation}
In particular, the association between the unobserved attribute and the score is transmitted partly through the observed proxy $X^\star$
\begin{equation}
\label{eq:projection_residual_proxy}
R_{\boldsymbol{D}}
=
\alpha_2
\left(
X^\star-
\mathbb E[X^\star\mid D_1,D_2]
\right).
\end{equation}

\paragraph{Correction using the observed attribute.} 

The Wasserstein barycentric correction targeting $D_1$ is therefore
\begin{equation}
\label{eq:proxy_correction}
\tilde{Z}_\lambda
=
\tilde{Z}-\lambda\alpha_1D_1
=
(1-\lambda)\alpha_1D_1
+\alpha_2X^\star,
\qquad \lambda\in[0,1]
\end{equation}
At $\lambda=0$, the original score is recovered; at $\lambda=1$, its conditional mean given $D_1$ is completely removed. This correction is implementable using only the observed variables.

\begin{proposition}[Disparity trajectories under an omitted sensitive attribute]\label{prop:trajectories-omitted}
Under the Gaussian model specified above, the marginal and joint squared Wasserstein barycentric disparity indices along the correction trajectory admit the following explicit expressions:
\begin{equation}
\label{eq:proxy_disparity_trajectories}
\begin{aligned}
\Ucal_1(\tilde{Z}_\lambda)
&=(1-\lambda)^2\alpha_1^2,
\\
\Ucal_2(\tilde{Z}_\lambda)
&=
\bigl[
(1-\lambda)\alpha_1\rho
+\alpha_2\tau
\bigr]^2,
\\
\Ucalj(\tilde{Z}_\lambda)
&=
(1-\lambda)^2\alpha_1^2
+
\frac{\alpha_2^2\tau^2}{1-\rho^2}.
\end{aligned}
\end{equation}
The externality on the unavailable sensitive attribute is
\begin{align}
\Delta  \tilde{\Ucal}_2(\lambda) =
\Ucal_2(\tilde{Z}_\lambda)-\Ucal_2(\tilde{Z})
\nonumber =
\underbrace{-\lambda(2-\lambda)\alpha_1^2\rho^2}_{<0}
-2\lambda\alpha_1\alpha_2\rho\tau.
\label{eq:proxy_externality}
\end{align}
The first term is non-positive and corresponds to the reduction in the association transmitted through $D_1$. The second term captures the interaction between the targeted correction and the residual
information about $D_2$ carried by $X^\star$.
\end{proposition}

\begin{proof}
For jointly Gaussian variables,
\begin{equation}
\mathbb E[\tilde{Z}_\lambda\mid D_1,D_2]
=(1-\lambda)\alpha_1D_1+\frac{\alpha_2\tau}{1-\rho^2}
(D_2-\rho D_1).
\end{equation}
Since $D_1$ is standard Gaussian, the variance of this conditional expectation is $\operatorname{Var}\bigl(\mathbb E[\tilde{Z}_\lambda\mid D_1]\bigr) =(1-\lambda)^2\alpha_1^2$ and the two Gaussian terms are independent
\[
D_1\perp(D_2-\rho D_1).
\]
Hence,
\[
\Ucalj(\tilde{Z}_\lambda)
=(1-\lambda)^2\alpha_1^2+
\frac{\alpha_2^2\tau^2}{1-\rho^2}.
\]
\end{proof}

\paragraph{Correction effect on the hidden disparity.}

The three trajectories in \eqref{eq:proxy_disparity_trajectories} have different interpretations. The targeted disparity $\Ucal_1$ decreases quadratically to zero. The joint disparity $\Ucalj$ is non-increasing throughout the correction, since its second term remains constant. By contrast, the non-target marginal disparity $\Ucal_2$ may increase or decrease, depending on the association between the proxy and the hidden attribute.

At full correction,
\[
\Ucal_1(\tilde{Z}_1)=0,
\qquad
\Ucal_2(\tilde{Z}_1)=\alpha_2^2\tau^2.
\]
Before correction, the hidden marginal disparity is
\begin{equation}
    \mathcal \Ucal_2(\tilde{Z})
=(\alpha_1\rho+\alpha_2\tau)^2.
\end{equation}
Consequently, full correction ($\lambda=1$) increases the hidden disparity if and only if
\begin{equation}
\label{eq:proxy_deterioration}
\Delta \tilde{\Ucal}_2(1)>0
\quad\Longleftrightarrow\quad
|\alpha_2\tau|
>
|\alpha_1\rho+\alpha_2\tau|.
\end{equation}
Thus, correcting the observed sensitive attribute can increase disparity associated with the unavailable attribute, even though the latter is never explicitly used to construct the score. This phenomenon results from the removal of a statistical compensation between the observed sensitive attribute and the proxy.

\paragraph{Partial identification when the hidden correlation is unknown.}

In practice, $\tau$ is unknown because $D_2$ is not observed. Even if $\rho, \alpha_1$, and $\alpha_2$ are known, the exact externality cannot generally be evaluated. The positive-semidefiniteness constraint, however, restricts the possible values of $\tau$ to
\[
|\tau|\leq\sqrt{1-\rho^2}.
\]
Since $\Delta \tilde{\Ucal}_2(\lambda)$ is affine in $\tau$, optimizing over this interval yields the following sharp bounds.

\begin{proposition}[Sharp bounds on the hidden externality]\label{prop:bounds-hidden}
Suppose that $\rho$, $\alpha_1$, and $\alpha_2$ are fixed and that the only unknown parameter is $\tau$, subject to $|\tau|\leq\sqrt{1-\rho^2}$. Then, for every $\lambda\in[0,1]$,
\begin{equation}\label{eq:interval-hidden}
\Delta \tilde{\Ucal}_2(\lambda) \in \left[\underline{\Delta}(\lambda), \overline{\Delta}(\lambda)\right],
\end{equation}
where,
\begin{equation}\label{eq:bounds-hidden}
\begin{aligned}
\underline{\Delta}(\lambda)
=&-\lambda(2-\lambda)\alpha_1^2\rho^2\
-2\lambda|\alpha_1\alpha_2\rho|
\sqrt{1-\rho^2},\\
\overline{\Delta}(\lambda)
=&-\lambda(2-\lambda)\alpha_1^2\rho^2\
+2\lambda|\alpha_1\alpha_2\rho|
\sqrt{1-\rho^2}.
\end{aligned}
\end{equation}
These bounds are attainable over the feasible parameter space.
\end{proposition}

\begin{proof}
For fixed $\lambda$ and $\rho$, $\Delta \tilde{\Ucal}_2(\lambda)$ is affine in $\tau$. Its minimum and maximum over the feasible interval are therefore attained at an endpoint, unless its coefficient of $\tau$ vanishes, in which case the externality is constant in $\tau$. Substituting $\tau=\pm\sqrt{1-\rho^2}$ and ordering the resulting values gives \eqref{eq:bounds-hidden}.
\end{proof}

\section{Derivation of Bounds for Categorical Attributes}

\subsection{Empirical Procedure for the Derivation of Bounds}
\label{app:subsec:empirical-bounds}

\begin{proposition}[\emph{Sharp empirical bounds from order statistics}]
\label{prop:sharp-hidden-sample}
Consider $n$ observations $(z_i,z_i',D_{1i})_{i=1}^n$, where
$z_i$ and $z_i'$ denote the original and corrected scores, respectively,
and $D_{1i}$ is the observed sensitive attribute. Suppose that the
number $m_a$ of individuals with hidden category $D_2=1$ is known
within each observed group $\{D_1=a\}$.

Define the set of feasible hidden-group assignments by
\[
\mathcal A
=
\left\{
h\in\{0,1\}^n:
\sum_{i:D_{1i}=a}h_i=m_a
\text{ for every }a
\right\}.
\]
Assume that $n$ is even and that
$\sum_a m_a=n/2$, so that each hidden group contains $n/2$
individuals.

For $h\in\mathcal A$ and $j\in\{0,1\}$, let
$z_{(k),j}(h)$ and $z'_{(k),j}(h)$ denote the $k$-th order
statistics of the original and corrected scores, respectively,
among individuals assigned to hidden group $D_2=j$.

Then, for every $h\in\mathcal A$,
\begin{equation}\label{eq:sharp-hidden-order}
\Delta \mathcal U_2(h)=\frac1{2n}\sum_{k=1}^{n/2}
\left[\bigl(z'_{(k),1}(h)-z'_{(k),0}(h)\bigr)^2-
\bigl(z_{(k),1}(h)-z_{(k),0}(h)\bigr)^2\right].
\end{equation}
The empirical identified interval is exactly
$[\min_{h\in\mathcal A}\Delta \mathcal U_2(h),
\max_{h\in\mathcal A}\Delta \mathcal U_2(h)]$. The proof is given in Appendix \ref{app:subsec:proof-prop-3-4}.
\end{proposition}


Proposition~\ref{prop:sharp-hidden-categorical} concerns the population and allows a probabilistic allocation of the hidden attribute. Proposition~\ref{prop:sharp-hidden-sample} instead concerns a calibrated empirical distribution with fixed hidden-group counts. An external proportion $q_a$ does not, by itself, determine the realized sample count $m_a$. In practice, exact enumeration is feasible for small samples, but its complexity grows as $\prod_a\binom{n_a}{m_a}$. For large datasets, a heuristic search provides only inner bounds; numerically sharp bounds require certified global optimization. 
For a hidden attribute with $L$ categories, the same approach extends to a partition into $L$ sub-measures and the identity
\[
\Ucal_2=\sum_{j<\ell}q_jq_\ell W_2^2(P_j,P_\ell).
\]


\subsection{Synthetic Data Example}


Consider two observed groups $D_1\in\{-1,1\}$, each containing four residual scores $W=(-1,0,0,1)$. Let the original score be $Z=\frac12D_1+W$, so that the barycentric correction gives $Z'=W$. Suppose only that half of the observations in each $D_1$ group have $D_2=1$; equivalently, $D_1\perp D_2$ in the empirical distribution. There are $\binom42^2=36$ admissible assignments. Table~\ref{tab:36-affectations} lists all 36 admissible assignments, where \texttt{0110} means that the second and third observations in the group are assigned to $D_2=1$; while Table~\ref{tab:externalites_36} summarizes the distribution of $\Delta \mathcal{U}_2$ across the 36 assignments of the unobserved attribute $D_2$ that satisfy the demographic constraints. The externality is negative in 50\% of assignments, zero in 44.4\%, and positive in only 5.6\%. Values range from $-0.250$ to $0.125$, indicating that the intervention's effect on disparity with respect to $D_2$ is more often non-positive than positive.

The exact numeration gives
\[
{-0.250\le \Delta \Ucal_2\le 0.125.}
\]
The upper bound is attained by assigning $D_2=1$ to the observations with $W=0$ in both groups for the 36th assignment: then, $\Ucal_2(Z)=0.125$ and $\Ucal_2(Z')=0.250$.
Conversely, the lower bound is attained by assigning $D_2=1$ to the observations with $W=\pm1$ when $D_1=-1$, and to the observations with $W=0$ when $D_1=1$ for the 1st assignment: then, $\Ucal_2(Z)=0.250$ and $\Ucal_2(Z')=0$. The observed law of $(D_1,Z)$ and the cross-group proportions of $(D_1,D_2)$ are identical in both cases.

\begin{table}[t]
\centering
\caption{The 36 admissible assignments $h \in \mathcal{A}$ of the unobserved attribute $D_2$ across two observed groups of four observations, ordered by $\Delta \Ucal_2$. $D_{2,1:4}$ denotes the values assigned to $D_2$ in the group $D_1=-1$, while $D_{2,5:8}$ denotes the values assigned to $D_2$ in the group $D_1=+1$. Lower and upper bounds  on $\Delta \Ucal_2$ are shown in bold.}
\label{tab:36-affectations}
\setlength{\tabcolsep}{2pt}
\begin{tabular}{rccrrr@{\hspace{0.7cm}}rccrrr}
\hline
\multicolumn{6}{c}{Assignments 1--18} & \multicolumn{6}{c}{Assignments 19--36}\\
\hline
No & $D_{2,1:4}$ & $D_{2,5:8}$ & $\Ucal_2(Z)$ & $\Ucal_2(Z')$ & $\Delta \Ucal_2$ & No & $D_{2,1:4}$ & $D_{2,5:8}$ & $\Ucal_2(Z)$ & $\Ucal_2(Z')$ & $\Delta \Ucal_2$ \\
\hline
\textbf{1} & \texttt{\textbf{0110}} & \texttt{\textbf{1001}} & \textbf{0.250} & \textbf{0.000} & \textbf{-0.250} & 19 & \texttt{0011} & \texttt{0011} & 0.250 & 0.250 & +0.000 \\
\textbf{2} & \texttt{\textbf{1001}} & \texttt{\textbf{0110}} & \textbf{0.250} & \textbf{0.000} & \textbf{-0.250} &
20 & \texttt{0011} & \texttt{0101} & 0.250 & 0.250 & +0.000 \\
3 & \texttt{0011} & \texttt{0110} & 0.250 & 0.125 & -0.125 & 21 & \texttt{0011} & \texttt{1001} & 0.125 & 0.125 & +0.000 \\
4 & \texttt{0011} & \texttt{1010} & 0.125 & 0.000 & -0.125 & 22 & \texttt{0101} & \texttt{0011} & 0.250 & 0.250 & +0.000 \\
5 & \texttt{0011} & \texttt{1100} & 0.125 & 0.000 & -0.125 & 23 & \texttt{0101} & \texttt{0101} & 0.250 & 0.250 & +0.000 \\
6 & \texttt{0101} & \texttt{0110} & 0.250 & 0.125 & -0.125 & 24 & \texttt{0101} & \texttt{1001} & 0.125 & 0.125 & +0.000 \\
7 & \texttt{0101} & \texttt{1010} & 0.125 & 0.000 & -0.125 & 25 & \texttt{0110} & \texttt{0011} & 0.125 & 0.125 & +0.000 \\
8 & \texttt{0101} & \texttt{1100} & 0.125 & 0.000 & -0.125 & 26 & \texttt{0110} & \texttt{0101} & 0.125 & 0.125 & +0.000 \\
9 & \texttt{0110} & \texttt{1010} & 0.250 & 0.125 & -0.125 & 27 & \texttt{1001} & \texttt{1010} & 0.125 & 0.125 & +0.000 \\
10 & \texttt{0110} & \texttt{1100} & 0.250 & 0.125 & -0.125 & 28 & \texttt{1001} & \texttt{1100} & 0.125 & 0.125 & +0.000 \\
11 & \texttt{1001} & \texttt{0011} & 0.250 & 0.125 & -0.125 & 29 & \texttt{1010} & \texttt{0110} & 0.125 & 0.125 & +0.000 \\
12 & \texttt{1001} & \texttt{0101} & 0.250 & 0.125 & -0.125 & 30 & \texttt{1010} & \texttt{1010} & 0.250 & 0.250 & +0.000 \\
13 & \texttt{1010} & \texttt{0011} & 0.125 & 0.000 & -0.125 & 31 & \texttt{1010} & \texttt{1100} & 0.250 & 0.250 & +0.000 \\
14 & \texttt{1010} & \texttt{0101} & 0.125 & 0.000 & -0.125 & 32 & \texttt{1100} & \texttt{0110} & 0.125 & 0.125 & +0.000 \\
15 & \texttt{1010} & \texttt{1001} & 0.250 & 0.125 & -0.125 & 33 & \texttt{1100} & \texttt{1010} & 0.250 & 0.250 & +0.000 \\
16 & \texttt{1100} & \texttt{0011} & 0.125 & 0.000 & -0.125 & 34 & \texttt{1100} & \texttt{1100} & 0.250 & 0.250 & +0.000 \\
17 & \texttt{1100} & \texttt{0101} & 0.125 & 0.000 & -0.125 & \textbf{35} & \texttt{\textbf{0110}} & \texttt{\textbf{0110}} & \textbf{0.125} & \textbf{0.250} & \textbf{+0.125} \\
18 & \texttt{1100} & \texttt{1001} &
0.250 & 0.125 & -0.125 &
\textbf{36} & \texttt{\textbf{1001}} & \texttt{\textbf{1001}} &
\textbf{0.125} & \textbf{0.250} & \textbf{+0.125} \\
\hline
\end{tabular}
\end{table}

\begin{table}[t]
\centering
\caption{Distribution of externalities compatible with the demographic constraints, across the 36 admissible assignments of the unobserved attribute $D_2$ from Table~\ref{tab:36-affectations}.}
\label{tab:externalites_36}
\begin{tabular}{rcc}
\hline
$\Delta \Ucal_2$ & Number of assignments & Proportion \\
\hline
$-0.250$ & 2  & $5.6\%$  \\
$-0.125$ & 16 & $44.4\%$ \\
$\phantom{-}0.000$ & 16 & $44.4\%$ \\
$\phantom{+}0.125$ & 2 & $5.6\%$ \\
\hline
{Total} & {36} & {100\%} \\
\hline
\end{tabular}
\end{table}

\section{Real-Data Experiments}

\subsection{Binary Classification}
\label{app:subsec:binary-classification}

\begin{table}[t]
\centering
\caption{\textbf{Changes in hidden-attribute unfairness, $\Delta \Ucal_h$, under pre-processing mitigation \textbf{DIR} and \textbf{RW}.} For each dataset and mitigated--hidden attribute pair, reports include the mean change (standard deviation in parentheses), minimum, maximum, and \% of runs with $\Delta \Ucal_h>0$ (indicating a deterioration in fairness), across 30 random seeds and four classification models (LR, RF, SVM, and DNN). $\rho_{1h}$ is the correlation between the mitigated and hidden attributes.}
\label{tab:delta_unf_dir_rw}
\small
\setlength{\tabcolsep}{4pt}
\begin{tabular}{ll l l r r r r r}
\toprule
\textbf{Mitigation} &
\textbf{Dataset} &
\textbf{Mitigated} &
\textbf{Hidden} &
$\boldsymbol{\rho_{1h}}$ &
\textbf{Mean $\Delta \Ucal_h$ (sd)} &
\textbf{Min $\Delta \Ucal_h$} &
\textbf{Max $\Delta \Ucal_h$} &
\textbf{\% $\Delta \Ucal_h > 0$} \\
\midrule
\multirow{10}{*}{DIR}
& Adult & Race & Sex & 0.101 & -0.026 (0.052) & -0.253 & 0.238 & 20.0 \\
& & Sex & Race & 0.101 & -0.026 (0.046) & -0.211 & 0.150 & 15.0 \\
& Compas & Race & Sex & 0.066 & -0.002 (0.017) & -0.069 & 0.050 & 34.2 \\
& & Sex & Race & 0.066 & -0.006 (0.013) & -0.046 & 0.034 & 25.8 \\
& Default & Age & Sex & -0.068 & -0.012 (0.024) & -0.064 & 0.105 & 20.8 \\
& & Sex & Age & -0.068 & -0.002 (0.040) & -0.177 & 0.217 & 40.8 \\
& Mep15 & Race & Sex & -0.017 & 0.012 (0.028) & -0.022 & 0.150 & 78.3 \\
& & Sex & Race & -0.017 & 0.000 (0.019) & -0.081 & 0.068 & 60.8 \\
& Mep16 & Race & Sex & -0.018 & -0.000 (0.026) & -0.108 & 0.113 & 50.8 \\
& & Sex & Race & -0.018 & 0.002 (0.017) & -0.084 & 0.091 & 56.7 \\
\midrule
\multirow{10}{*}{RW}
& Adult & Race & Sex & 0.101 & -0.131 (0.095) & -0.306 & 0.249 & 6.7 \\
& & Sex & Race & 0.101 & -0.090 (0.082) & -0.242 & 0.130 & 15.0 \\
& Compas & Race & Sex & 0.066 & -0.053 (0.029) & -0.116 & 0.017 & 4.2 \\
& & Sex & Race & 0.066 & -0.064 (0.038) & -0.139 & 0.027 & 10.0 \\
& Default & Age & Sex & -0.068 & -0.022 (0.019) & -0.064 & 0.049 & 12.5 \\
& & Sex & Age & -0.068 & -0.025 (0.043) & -0.100 & 0.200 & 15.0 \\
& Mep15 & Race & Sex & -0.017 & -0.033 (0.032) & -0.113 & 0.107 & 15.0 \\
& & Sex & Race & -0.017 & -0.019 (0.025) & -0.084 & 0.067 & 18.3 \\
& Mep16 & Race & Sex & -0.018 & -0.031 (0.034) & -0.123 & 0.074 & 15.8 \\
& & Sex & Race & -0.018 & -0.023 (0.026) & -0.075 & 0.083 & 16.7 \\
\bottomrule
\end{tabular}
\end{table}

Comparing the three mitigation methods using Table~\ref{tab:delta-unf-summary-two-attr-binary} and Table~\ref{tab:delta_unf_dir_rw} reveals substantial differences in their cross-attribute effects. While \textbf{EquiPy} and \textbf{DIR} generally produce relatively small changes in hidden-attribute unfairness, \textbf{RW} consistently yields negative mean changes across all datasets and attribute pairs, indicating systematic reductions in hidden-attribute unfairness. This effect is particularly pronounced for Adult and COMPAS. For Default and Mep15, \textbf{EquiPy} and \textbf{DIR} exhibit positive mean changes in some scenarios, suggesting that mitigating one sensitive attribute does not necessarily improve fairness with respect to another. Overall, these results highlight the dependence of cross-attribute externalities on both the mitigation method and the underlying dataset.

\begin{figure}[ht]
\centering\includegraphics[width=.8\linewidth]{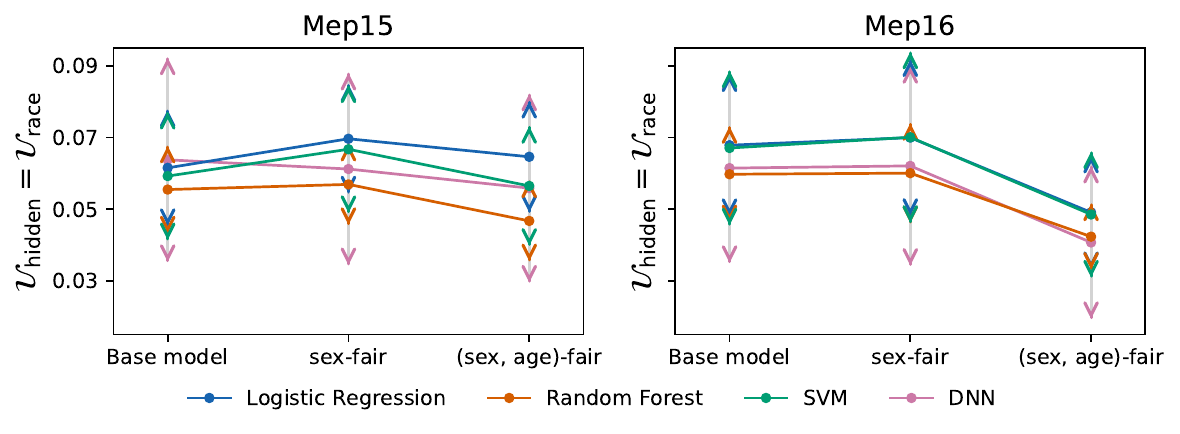}
\caption{\textbf{Model-dependent mitigation externalities.} The figure compares hidden-attribute unfairness across the baseline model, the intermediate mitigation step, and the subsequent joint correction for the LR, SVM, RF, and DNN classification models. The direction of the change in unfairness following each correction depends on the model. For Mep15, an individual mitigation step may either increase (LR, RF, SVM) or decrease (DNN) hidden-attribute unfairness, whereas for Mep16, it consistently increases it. For both datasets, the subsequent joint correction reduces hidden-attribute unfairness, although it sometimes remains above the baseline level (Mep15 with LR). These results highlight the model- and dataset- dependent nature of fairness externalities under sequential mitigation.}
\label{fig:binary-classif-u3-models}
\end{figure}

Figure~\ref{fig:binary-classif-u3-models} shows how mitigating unfairness with respect to sex, alone or jointly with age, affects race-related unfairness across four classification models on Mep15 and Mep16 datasets. The effects of sex-based mitigation vary across models: on Mep15, race unfairness increases most notably for LR and SVM, while changes are smaller for RF and DNN. For Mep16, unfairness changes only marginally during the first mitigation step, whereas the magnitude of the changes increases during the subsequent intervention across all models. DNN exhibits the largest standard deviations, indicating greater variability in hidden-attribute unfairness across runs. Overall, these results highlight the model- and dataset-dependent nature of cross-attribute externalities and show that sequential correction can produce non-monotonic changes in hidden-attribute unfairness across all models.

\begin{figure}[t]
    \centering
    \begin{subfigure}[t]{0.48\textwidth}
        \centering
        \includegraphics[width=\linewidth]
        {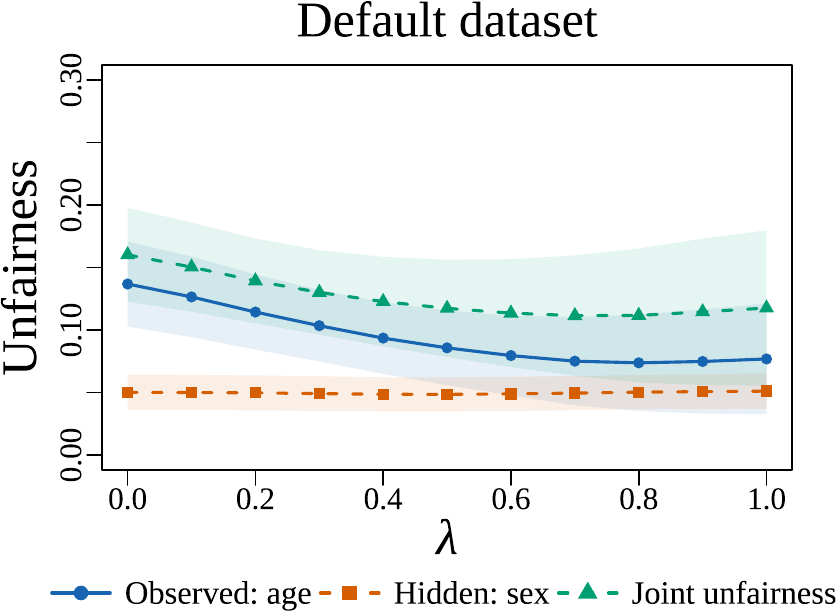}
    \end{subfigure}
    \hfill
    \begin{subfigure}[t]{0.48\textwidth}
        \centering
        \includegraphics[width=\linewidth]
        {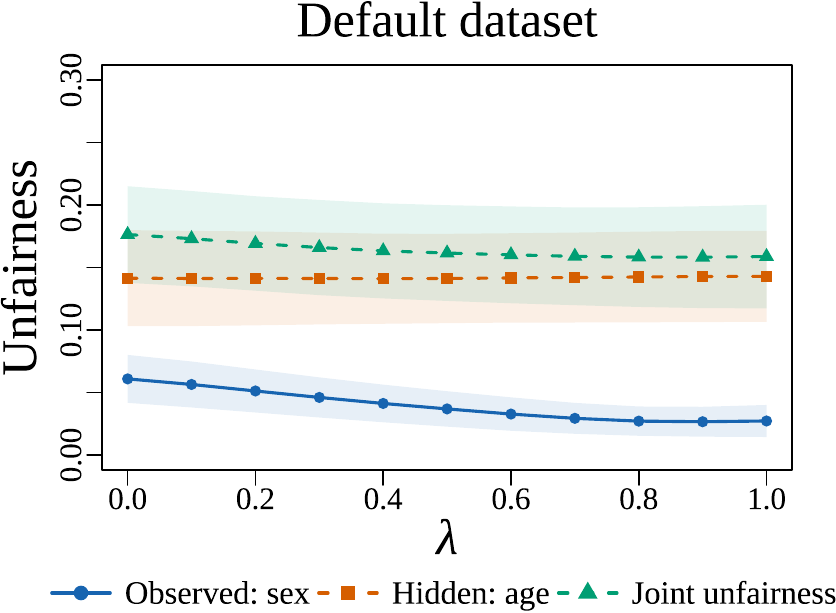}
        \label{fig:z_lambda_acs_age_sex}
    \end{subfigure}
    \vspace{0.5em}
    \begin{subfigure}[t]{0.48\textwidth}
        \centering
        \includegraphics[width=\linewidth]
        {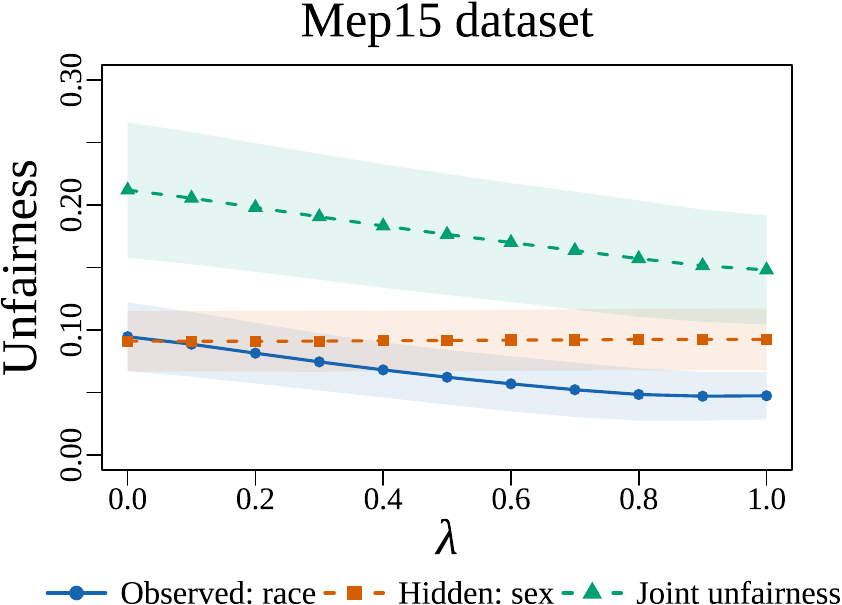}
        \label{fig:z_lambda_crime_poverty_age}
    \end{subfigure}
    \hfill
    \begin{subfigure}[t]{0.48\textwidth}
        \centering
        \includegraphics[width=\linewidth]
        {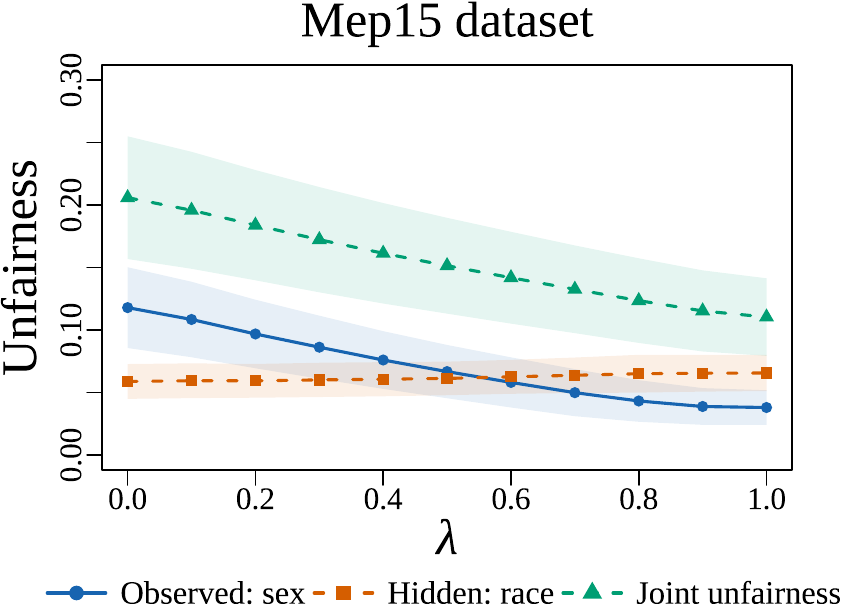}
        \label{fig:unfairness-lambda-default-mep}
    \end{subfigure}
    \caption{\textbf{Unfairness as a function of the mitigation parameter $\lambda$.} Each panel reports the mean observed, hidden-attribute, and joint unfairness across random seeds and classification models (LR, SVM, RF, and DNN) for the indicated dataset and attribute configuration. Solid, dashed, and dotted lines distinguish observed, hidden-attribute, and joint unfairness, respectively; shaded bands represent one standard deviation around the mean. For Mep15, when mitigating sex (bottom-right panel), race-related unfairness increases as \(\lambda\) grows, while joint unfairness decreases markedly. This illustrates that improvements in joint fairness can coexist with adverse externalities on a hidden attribute, highlighting the importance of examining both marginal and joint indices.}
\end{figure}

\begin{table}[t]
\centering
\caption{\textbf{Changes in hidden-attribute unfairness under Wasserstein barycenter mitigation (\textbf{EquiPy}) with two observed attributes.} For each dataset with three sensitive attributes (Mep15 and Mep16), we consider pairs of mitigated attributes and a hidden attribute $h$. Results are aggregated over 30 random seeds and four classification models (LR, SVM, RF, and DNN) out-of-sample. Reports include the mean change in unfairness (standard deviation in parentheses), minimum, maximum, and \% of runs with $\Delta \Ucal_h>0$ (indicating increased hidden-attribute unfairness). Here, $\rho_{1h}$ and $\rho_{2h}$ denote the correlations between the first and second mitigated attributes, respectively, and the hidden attribute $h$.}
\label{tab:delta-unf-summary-three-attr-binary}
\small
\setlength{\tabcolsep}{5pt}
\begin{tabular}{ll l r r r r r r}
\toprule
\textbf{Dataset} &
\textbf{Mitigated} &
\textbf{Hidden} &
$\boldsymbol{\rho_{1h}}$ & 
$\boldsymbol{\rho_{2h}}$ &
\textbf{Mean $\Delta \Ucal_h$} &
\textbf{Min $\Delta \Ucal_h$} &
\textbf{Max $\Delta \Ucal_h$} &
\textbf{\% $\Delta \Ucal_h$ > 0} \\
\midrule
Mep15 & Age/Sex & Race & -0.013 & -0.016 & $-0.004$ (0.014) & -0.053 & 0.036 & 37.1 \\
\midrule
Mep16 & Age/Sex & Race & -0.046 & -0.017 & $-0.018$ (0.014) & -0.056 & 0.017 & 8.3 \\
\bottomrule
\end{tabular}
\end{table}

Table~\ref{tab:delta-unf-summary-three-attr-binary} shows that simultaneously mitigating age and sex generally reduces race-related unfairness on average, but the proportion of runs with increased unfairness reveals a more nuanced pattern. For Mep15, the mean change is close to zero ($-0.004$), yet $\Delta \mathcal{U}_h>0$ in 37.1\% of runs. For Mep16, the mean reduction is larger ($-0.018$), and race-related unfairness increases in a smaller proportion of runs (only 8.3\%).

\subsection{Regression}

\begin{table}[t]
\centering
\caption{\textbf{Changes in hidden-attribute unfairness, $\Delta \Ucal_h$, under Wasserstein barycenter mitigation (\textbf{EquiPy})}. For each dataset and mitigated--hidden attribute pair, $\Delta \Ucal_h$ is summarized across 30 random seeds and the regression models (GLM, RF, and NN) considered. The table reports the mean change (standard deviation in parentheses), minimum, maximum, and percentage of runs with $\Delta \Ucal_h>0$, and $\rho_{1h}$ denotes the correlation between the mitigated and hidden attributes. Positive values indicate an increase in hidden-attribute unfairness.}
\label{tab:delta_uh_summary_sens2}
\small
\setlength{\tabcolsep}{5pt}
\begin{tabular}{ll l r r r r r r}
\toprule
\textbf{Dataset} &
\textbf{Mitigated} &
\textbf{Hidden} &
$\boldsymbol{\rho_{1h}}$ & 
\textbf{Mean $\Delta \Ucal_h$ (sd)} &
\textbf{Min $\Delta \Ucal_h$} &
\textbf{Max $\Delta \Ucal_h$} &
\textbf{\% $\Delta \Ucal_h$ > 0} \\
\midrule
Law School & Sex & Race & 0.077 & -0.016 (0.013) & -0.053 & 0.007 & 7.8\\
\midrule
& Age & Sex & 0.003 & -0.033 (0.0132) & -0.650 & 0.300 & 31.1 \\
\midrule
Communities and Crime & Age & Race &0.085 & -0.004 (0.009) & -0.024 & 0.036 & 22.2 \\%
& Poverty & Age & 0.138 & \bf0.002 (0.031) & -0.070 & 0.087 & 54.4 \\
& Race & Poverty & -0.336 & -0.181 (0.050) & -0.314 & -0.041 & 0.0 \\
\midrule
ACSIncome (PA) & Sex & Race &  0.022 & -0.006 (0.043) & -0.188 & 0.176 & 41.1 \\
& Sex & Age & 0.012 & \bf0.006 (0.039) & -0.083 & 0.138 & 60.0 \\
& Age & Race & 0.090 & -0.498 (0.349) & -1.157 & 0.182 & 5.6 \\
\bottomrule
\end{tabular}
\end{table}

\begin{figure}[t]
    \centering
    \begin{subfigure}[t]{0.48\textwidth}
        \centering
        \includegraphics[width=\linewidth]
        {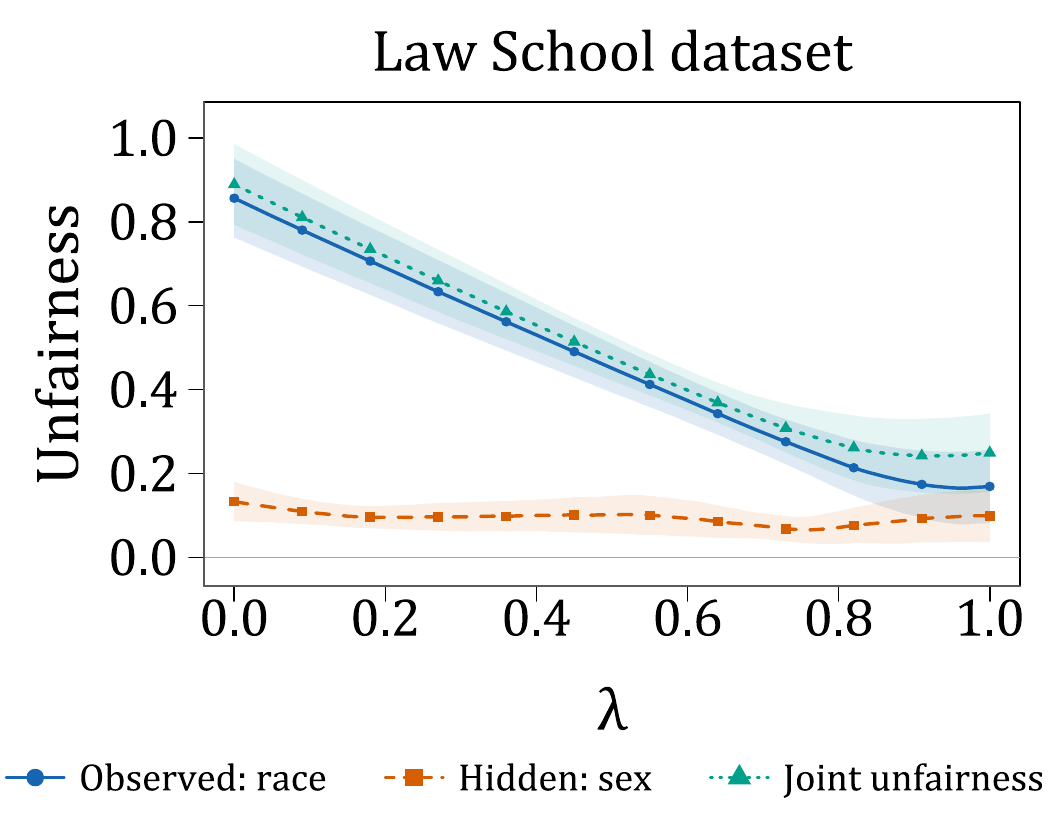}
    \end{subfigure}
    \hfill
    \begin{subfigure}[t]{0.48\textwidth}
        \centering
        \includegraphics[width=\linewidth]
        {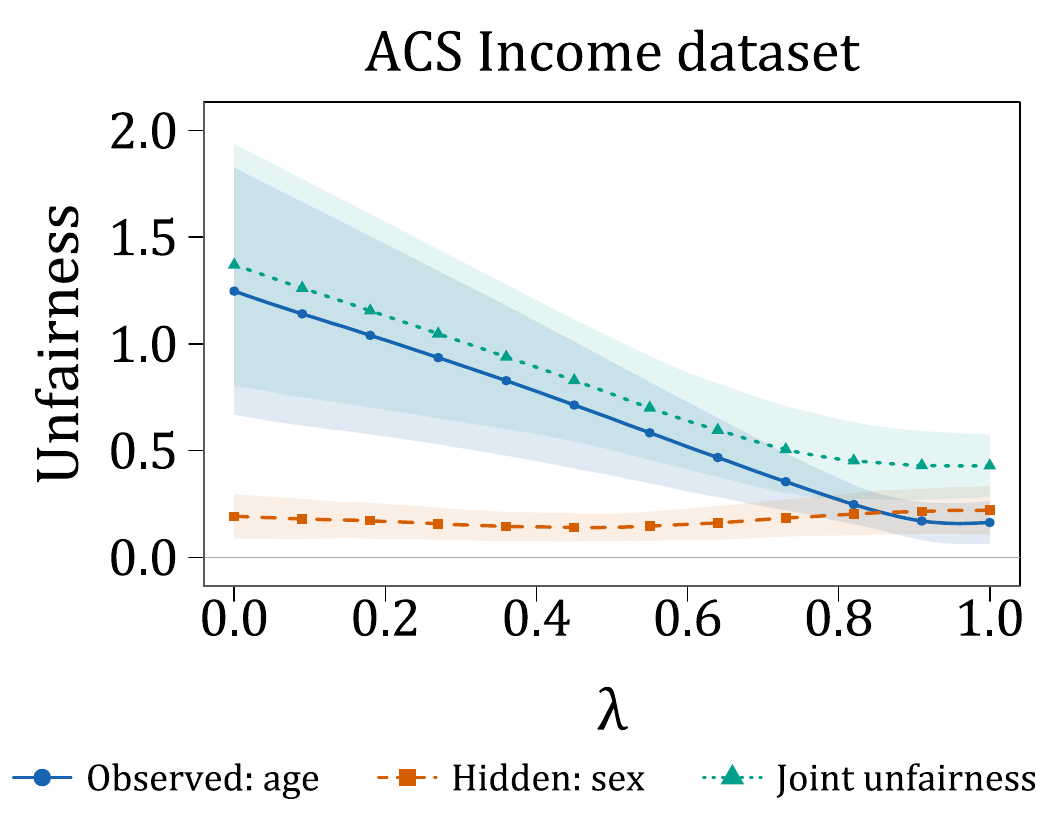}
    \end{subfigure}
    \vspace{0.5em}
    \begin{subfigure}[t]{0.48\textwidth}
        \centering
        \includegraphics[width=\linewidth]
        {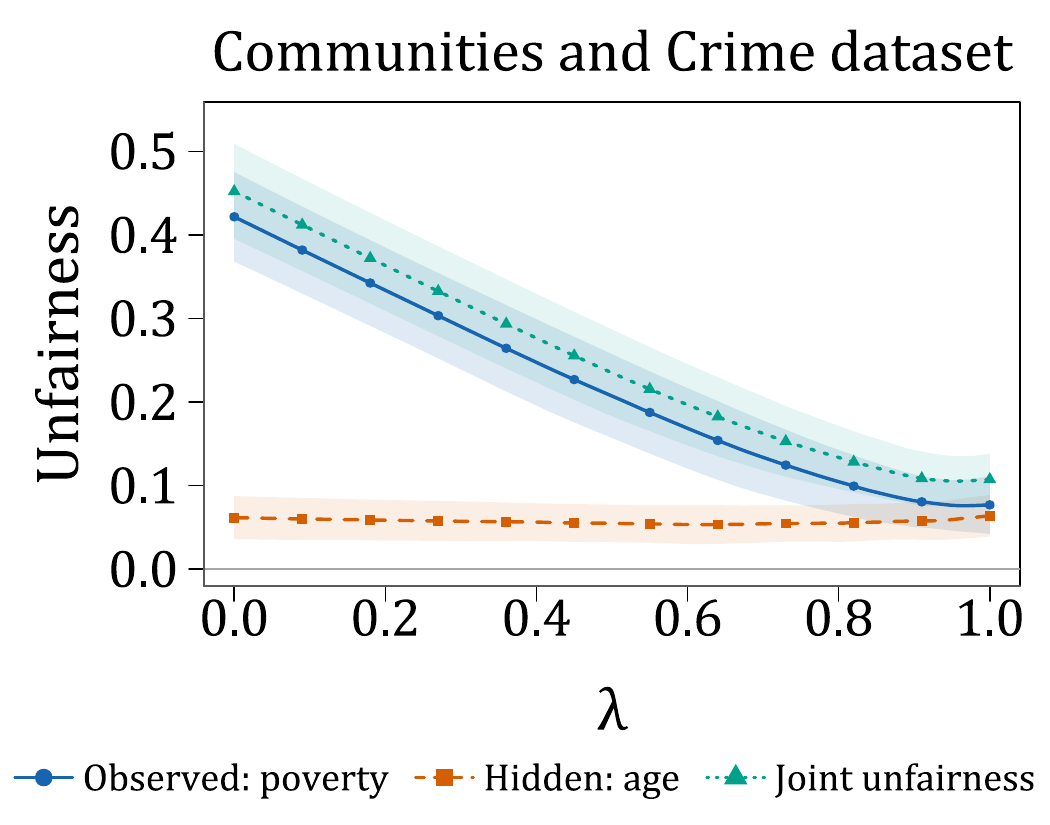}
    \end{subfigure}
    \hfill
    \begin{subfigure}[t]{0.48\textwidth}
        \centering
        \includegraphics[width=\linewidth]
        {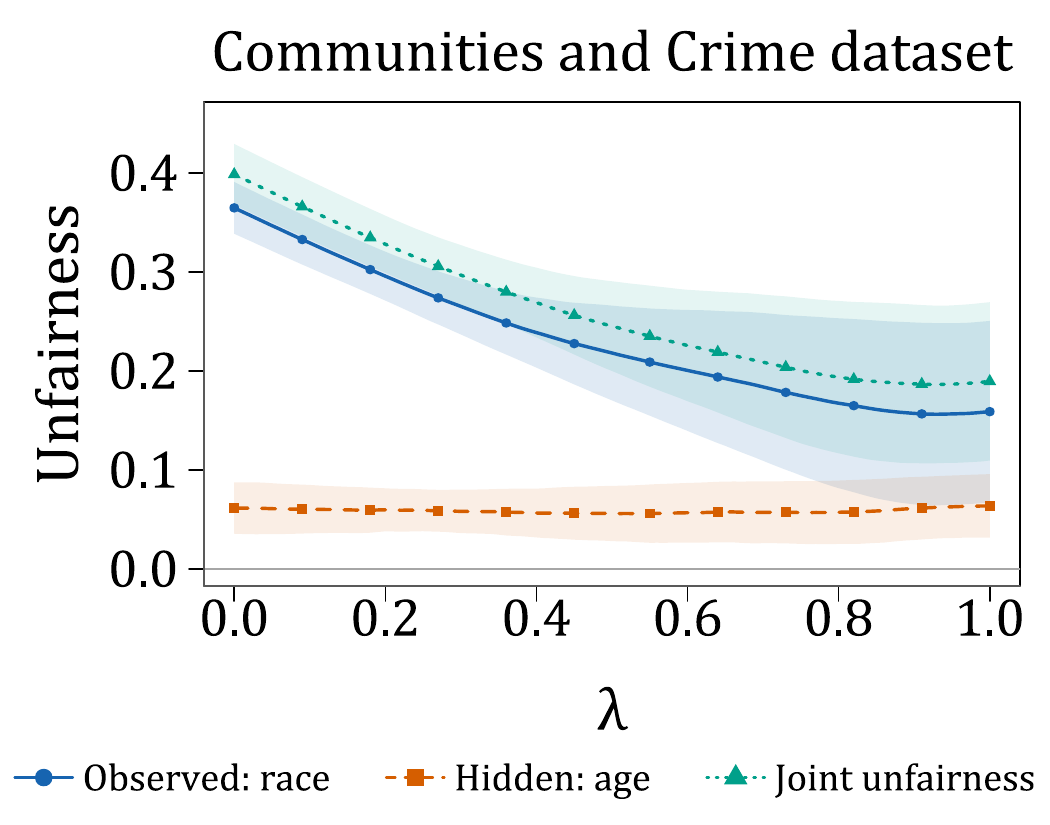}
    \end{subfigure}
    \caption{\textbf{Unfairness as a function of the mitigation parameter $\lambda$.} Each panel reports the mean observed, hidden-attribute, and joint unfairness across random seeds and regression models (GLM, RF, and NN) for the indicated dataset and attribute configuration out-of-sample. Solid, dashed, and dotted lines distinguish observed, hidden-attribute, and joint unfairness, respectively; shaded bands represent one standard deviation around the mean.}
    \label{fig:z_lambda_unfairness_comparison}
\end{figure}

While Table \ref{tab:delta_uh_summary_sens1} summarizes the overall change in hidden-attribute unfairness, Figure \ref{fig:z_lambda_unfairness_comparison} show whether that change is stable across $\lambda$. In the Law School experiment, when mitigating race while sex remains hidden, the hidden-attribute unfairness fluctuates with $\lambda$. This illustrates that improvements in the targeted and joint criteria need not translate into a monotonic reduction in hidden-attribute unfairness. In ACSIncome, mitigating age while sex remains hidden is accompanied by an increase in hidden-attribute unfairness around $\lambda=0.6$. This pattern is consistent with the positive mean changes reported for the age-sex configurations in Table \ref{tab:delta_uh_summary_sens1}. In the Communities and Crime datasets, when poverty or race is mitigated while age remains hidden, hidden-attribute unfairness exhibits a slight increase at higher values of $\lambda$ (from approximately $\lambda=0.8$ onward). 

\begin{figure}
    \centering
    \includegraphics[width=0.45\linewidth]{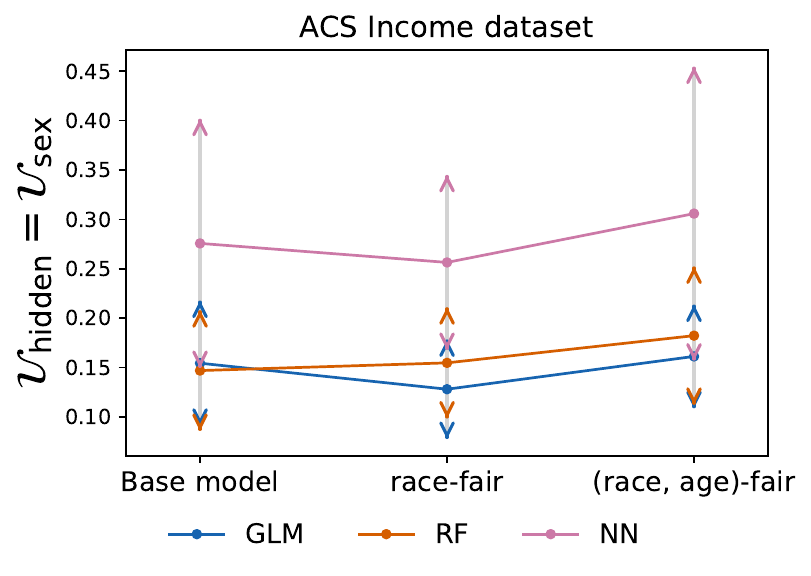}\includegraphics[width=0.45\linewidth]{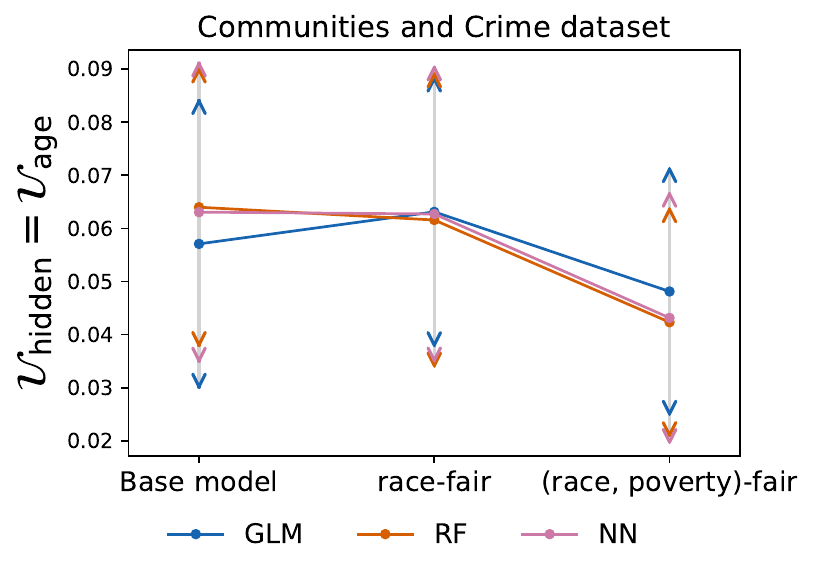}
    \caption{\textbf{Model-dependent mitigation externalities.} The figure compares hidden-attribute unfairness across the baseline model, the intermediate mitigation step, and the subsequent joint correction for the GLM, RF, and NN predictive models. The direction of the change in unfairness following each correction depends on the model: an individual mitigation step may increase or decrease hidden-attribute unfairness, and the subsequent joint correction may further increase it or reverse the intermediate improvement. These results highlight the model dependence of fairness externalities under sequential mitigation.}
    \label{fig:model_dependent}
\end{figure}

Figure~\ref{fig:model_dependent} illustrates that the effects of mitigation on hidden-attribute unfairness depend on the predictive model. For ACSIncome, the change in unfairness with respect to hidden age does not follow a uniform pattern across the GLM, RF, and NN models. Depending on the model, the first mitigation step can increase hidden-attribute unfairness, with the subsequent joint correction producing a further increase. For other models, including the NN and GLM in the configurations shown, the first correction instead reduces unfairness, before the joint correction increases it again. In Communities and Crime, the GLM exhibits an increase in hidden-age unfairness after the first mitigation step, whereas, both the RF and NN models show a decrease in unfairness after the first correction, with further reductions following joint mitigation. These contrasting trajectories indicate that the fairness externalities of mitigation are model-dependent: the same mitigation procedure can produce different intermediate and cumulative effects depending on the predictive model to which it is applied.

\end{document}